\documentclass[journal]{IEEEtai}

\usepackage[colorlinks,urlcolor=blue,linkcolor=blue,citecolor=blue]{hyperref}

\usepackage{color,array}

\usepackage{graphicx}
\usepackage{amsmath,amsfonts}
\usepackage{algorithmic}
\usepackage{algorithm}
\usepackage{array}
\usepackage[caption=false,font=normalsize,labelfont=sf,textfont=sf]{subfig}
\usepackage{textcomp}
\usepackage{stfloats}
\usepackage{url}
\usepackage{verbatim}
\usepackage{graphicx}
\usepackage{cite}
\usepackage{booktabs}
\usepackage{multirow}
\usepackage{cite} 
\usepackage{longtable}
\usepackage[T1]{fontenc}
\usepackage{amsmath}

\begin{document}

\title{Deep Learning Methods for Small Molecule Drug Discovery: A Survey}

\author{Wenhao Hu$^*$, Yingying Liu$^*$, Xuanyu Chen, Wenhao Chai, Hangyue Chen, Hongwei Wang$^\dagger$,~\IEEEmembership{Member,~IEEE} and Gaoang Wang$^\dagger$,~\IEEEmembership{Member,~IEEE}
\thanks{$^*$ Equal contribution.}
\thanks{$^\dagger$ Corresponding author: Hongwei Wang, and Gaoang Wang.}
\thanks{Wenhao Hu, Yingying Liu, Xuanyu Chen, and Wenhao Chai are with the Zhejiang University-University of Illinois at Urbana-Champaign Institute, Zhejiang University, China (e-mail: wenhao.21@intl.zju.edu.cn, yingying.19@intl.zju.edu.cn, xuanyu.19@intl.zju.edu.cn, wenhaochai.19@intl.zju.edu.cn).}
\thanks{Hongwei Wang, and Gaoang Wang are with the Zhejiang University-University of Illinois at Urbana-Champaign Institute, and College of Computer Science and Technology, Zhejiang University, China (e-mail: hongweiwang@intl.zju.edu.cn, gaoangwang@intl.zju.edu.cn).}
\thanks{Hangyue Chen is with the Hangzhou Dianzi University, China (e-mail: chy@hdu.edu.cn).}
}


\maketitle

\begin{abstract}
With the development of computer-assisted techniques, research communities including biochemistry and deep learning have been devoted into the drug discovery field for over a decade. Various applications of deep learning have drawn great attention in drug discovery, such as molecule generation, molecular property prediction, retrosynthesis prediction, and reaction prediction. While most existing surveys only focus on one of the applications, limiting the view of researchers in the community. In this paper, we present a comprehensive review on the aforementioned four aspects, and discuss the relationships among different applications. The latest literature and classical benchmarks are presented for better understanding the development of variety of approaches.

We commence by summarizing the molecule representation format in these works, followed by an introduction of recent proposed approaches for each of the four tasks. Furthermore, we review a variety of commonly used datasets and evaluation metrics and compare the performance of deep learning-based models. Finally, we conclude by identifying remaining challenges and discussing the future trend for deep learning methods in drug discovery.

\end{abstract}

\begin{IEEEImpStatement}
As artificial intelligence continues to evolve, an increasing number of deep learning algorithms are proposed to utilize extensive experimental data to accelerate the time-consuming and costly procedure for drug discovery. Unlike previous surveys that tend to focus on a single aspect of drug discovery, we offer a more comprehensive survey that encompasses four areas of drug discovery utilizing deep learning techniques, including molecule generation, molecular property prediction, retrosynthesis, and reaction prediction. Before introducing the main methods, we start with molecular data representation, which is also crucial for deep learning algorithms. Moreover, we present the recent benchmarks for deep learning-based drug discovery, which cover 9 widely used databases. 
\end{IEEEImpStatement}


\begin{figure}[t]
     \centering
     \includegraphics[width=3.5 in]{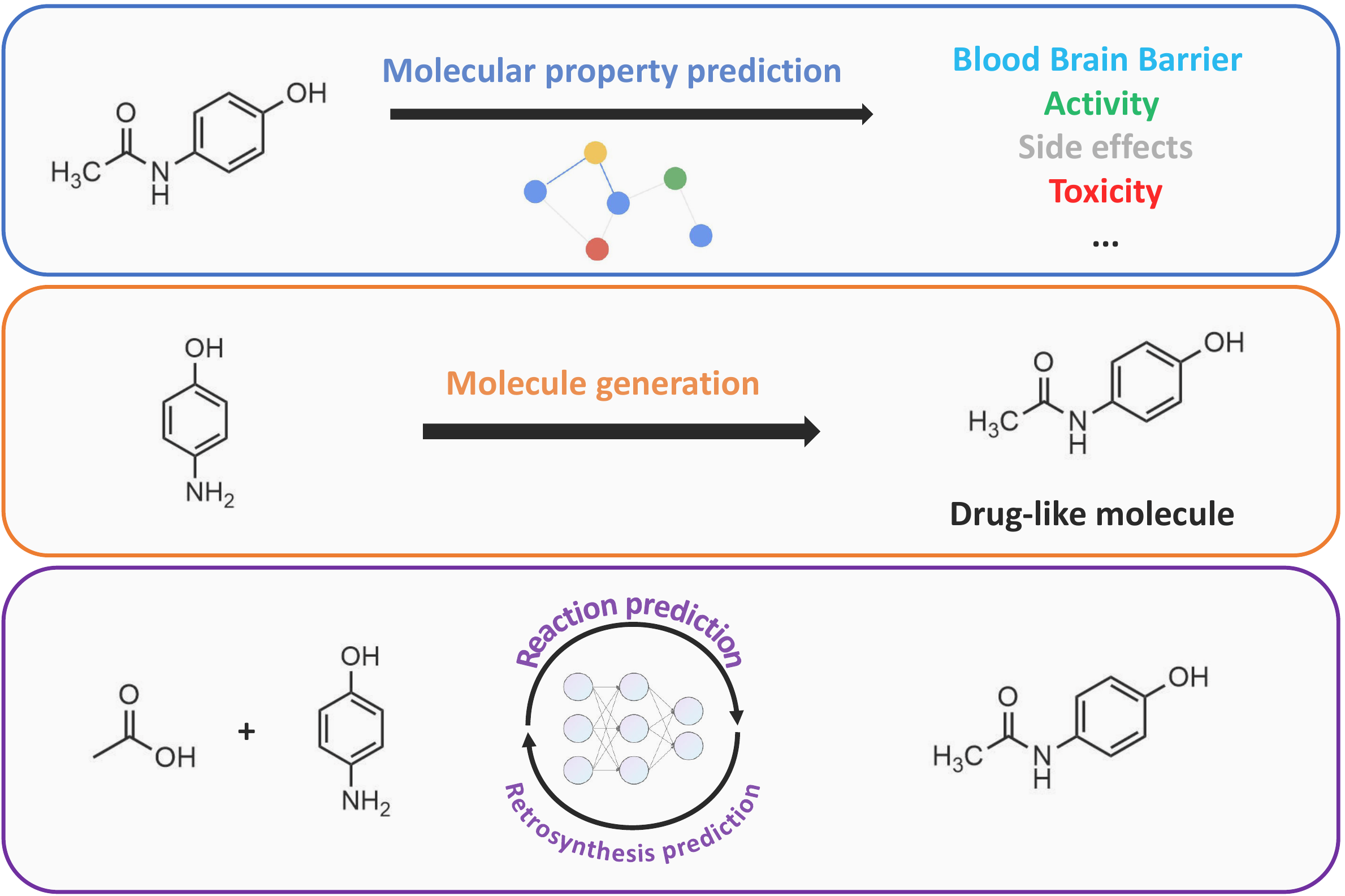}
     \vspace{-15pt}
     \caption{Framework of deep learning methods for drug discovery.}
     \label{fig:framework}
     \vspace{-13pt}
\end{figure}

\section{Introduction}

\IEEEPARstart{D}{rug} discovery is a very long process that involves extensive experiments and different stages. It may take more than ten years and cost around 2.6 billion dollars to discover a new drug \cite{deng2022artificial}. The number of the existing molecules has reached more than 130 million, which leads to the fact that the searching space for effective drug molecules is pretty significant \cite{schwaller2019molecular}. With the advent of artificial intelligence, more and more algorithms and methods are proposed to utilize the extensive experimental data from chemical labs to accelerate the time-consuming and costly procedure for drug discovery. 
Deep Learning can be applied to drug discovery in different stages. Among them, there are several vital aspects where the applications hugely help improve performance. In this survey, we focus on the following applications:
1) molecule generation, 2) molecular property prediction, 3) retrosynthesis, and 4) reaction prediction, as shown in Fig.~\ref{fig:framework}.

\textbf{Molecule generation} is the basis of rational drug design and the starting point for obtaining new drug molecules. The core problem is to obtain candidate molecules that satisfy specific properties and druggability from the huge chemical space. Mathematically, molecule generation is usually defined as an optimization problem in which the desired molecular properties are predicted by oracle functions. First, molecular characteristics are learned from a large set of molecules. Then, novel candidates are obtained from the learned distribution with the restriction of predefined oracle functions. In the early stage of molecule generation, molecules with novel structures can be constructed by combining fragments of existing compounds \cite{kumar2012fragment,sheng2013fragment}. This requires assistance of chemical experiment which is relatively inefficient. Optimization algorithms such as genetic algorithms \cite{sheridan1995using} are also considered for solving the molecule generation problem. Based on evolution principles, genetic algorithms obtain molecules with specific properties by changing parameters of a population. 
While the early approaches rely on stochastic steps that struggle to capture the constraints of chemical design \cite{schwalbe2020generative}. Deep learning brings the ability to learn from vast amounts of data and the potential for drug design beyond chemical intuition, narrowing the $10^{60}$ size chemical space to dozens of valid candidate compounds.

\textbf{Molecular property prediction} is a crucial part of drug development, and its primary purpose is to predict the drug activity, absorption, toxicity, and other properties of candidate compounds. Molecular property prediction has two main applications: quantitative structure-property relationship (QSPR) and virtual screening. QSPR models are usually defined as regression or classification models which predict property by establishing relationships between molecule structure and property. Early machine learning methods such as random forest \cite{svetnik2003random}, k nearest-neighbor \cite{zhang2008qsar}, support vector machine~\cite{frohlich2006kernel} have been used for QSPR problems. These methods still perform better than modern deep learning methods for some specific problem instances~\cite{chen2019deep,borisov2021deep,li2021machine,bomane2019paclitaxel,grinsztajn2022tree}. For example, algorithms using extremely randomized trees achieve better performance than deep learning methods on small datasets for organic molecular energy prediction problems~\cite{paul2019property}.

The purpose of virtual screening is to search the molecule library in order to identify those drug ligand structures which are most likely to bind to a target protein \cite{rester2008virtuality,rollinger2008virtual}. Traditional virtual screening methods need to synthesize many compounds for biological experiments, and the whole process has a high cost, long cycle, and low success rate. The virtual screening of drugs through deep learning is expected to replace traditional activity screening methods, speed up intermediate steps, and significantly reduce costs. For instance, a  particularly successful application is deep learning-based score function that can predict the affinities of protein-bound ligands with high accuracy \cite{ghislat2021recent}. More specifically, AtomNet incorporates the 3D structural features of the protein–ligand complexes to the CNNs \cite{wallach2015atomnet}. 
Feng et al. \cite{feng2018padme} combine the fringerprint and graph representations as ligand features. This model showes better performance on the Davis \cite{davis2011comprehensive}, Metz \cite{metz2011navigating}, and KIBA \cite{tang2014making} benchmark datasets. Gao et al. \cite{gao2018interpretable} use the LSTM recurrent neural networks for the protein sequences and the GCN for ligand structures.

\textbf{Retrosynthesis analysis} and \textbf{reaction prediction} are the critical links in drug synthesis. Reaction prediction explores possible product molecules synthesized by given reactants~\cite{yang2019molecular}. Contrary to reaction prediction, retrosynthesis is a reverse extrapolation from the product molecule to accessible starting materials \cite{dong2022deep}. 
Early computer-aided synthesis planning primarily relied on human-encoded reaction rules, which can be only applied to specific reactions \cite{cook2012computer}. With neural networks, the reaction patterns of molecules and the breaking rules of chemical bonds are learned from the existing massive chemical reaction data. These rules are applied to the retrosynthesis analysis and reaction prediction of drug molecules, which can significantly speed up the synthesis of new drugs.

Many existing surveys focus on drug discovery in the AI era. Lavecchia \cite{lavecchia2015machine} analyzes the application of machine learning techniques in ligand-based virtual screening. Gawehn et al.~\cite{gawehn2016deep} use several classic deep neural networks as examples to introduce the potential application fields. Some present how machine learning takes advantage of big data to improve the traditional method and the applications in various stages of drug discovery \cite{gupta2021artificial}. While those works focus on general aspects of drug discovery, other surveys emphasize one specific task and analyze state-of-the-art (SOTA) models for the particular application. For property prediction, some papers review the main methods, introduce several fundamental machine learning techniques, and analyze the performance from the data side \cite{nigam2021assigning,wieder2020compact}. For molecular design, Tong et al. \cite{tong2021generative} summarize generative models and list several tools which use those models, while Elton et al. \cite{elton2019deep} focus on the generation of lead molecules, classifying the models and comparing their performances. For the synthesis procedure, some surveys discuss the dataset, models, and tools for retrosynthesis and reaction prediction and point out the potential development directions \cite{dong2022deep,struble2020current}.

To the best of our knowledge, our survey first integrates the four applications mentioned above. The main contributions are summarized as follows:

\begin{itemize}
    \item{We summarize the latest literature and classical benchmarks in four applications and categorize them according to molecule representations.}
    \item{We enumerate commonly used datasets and evaluation metrics and compare the results of existing models.}
    \item{We propose the problems of existing methods and analyze the future development direction.}
\end{itemize}

Following the three main components of deep learning: representation, model, and data \cite{dobbelaere2021machine}, our organization for the paper will be as follows: we first briefly introduce the formats used in deep learning to represent the molecular structure in Section~\ref{sec:molecule representation}, including fingerprints, simplified molecular input line entry system (SMILES) and molecular graphs. 

After that, molecule generation, molecular property prediction, retrosynthesis and reaction prediction are introduced in Section~\ref{sec:Molecule Generation}, Section~\ref{sec:Molecular Property Prediction}, Section~\ref{sec:Retrosynthesis}, and Section~\ref{sec:Reaction Prediction}, respectively.
We also provide standard datasets, existing benchmark platforms, and usual evaluation metrics for all four tasks in Section~\ref{sec:datasets}. Finally, we discuss the challenges and possible development directions in Section~\ref{sec:discussion}.
\section{Molecule Representations}
\label{sec:molecule representation}

\begin{figure*}[t]
     \centering
     \includegraphics[width=0.7\linewidth]{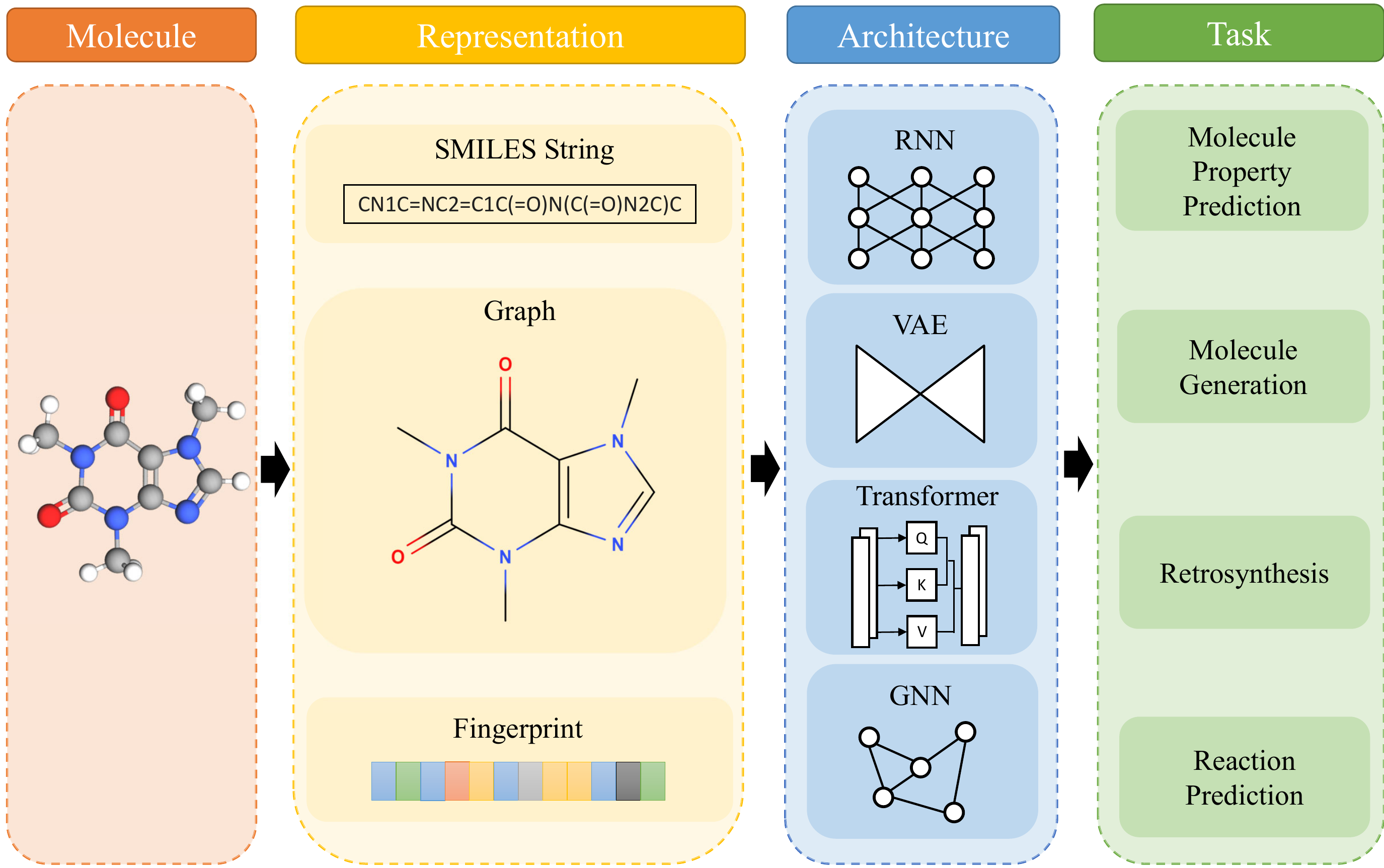}
     \vspace{-2pt}
     \caption{Taxonomy for deep neural network methods for drug discovery.}
     \label{fig:structure}
     \vspace{-10pt}
\end{figure*}

In order to convert the molecular structure to computer-readable information, there are several ways to represent the molecules. This section focuses on three common representations: fingerprints, SMILES string, and molecular graphs. After the introduction of the data representations, we show the taxonomy of the deep learning methods for drug discovery according to different representations in Fig.~\ref{fig:structure}.

\subsection{Fingerprints}
Molecular fingerprint is a bit string that encodes the structural information of a molecule. Two types of fingerprints are widely used for molecule representation, \textit{i.e.}, key-based fingerprints and hash fingerprints.

Key-based fingerprints, including molecular ACCess system (MACCS) \cite{durant2002reoptimization}, and PubChem fingerprint \cite{pubchem}, have a predefined fragment library so that each molecule can be encoded into a binary bit stream according to its substructure. The MACCS contains 166 predefined fragments and the fragment keys are implemented in many cheminformatics software packages, including RDKit \cite{RDKit}, OpenBabel \cite{o2011open}, CDK \cite{CDK}. The PubChem fingerprint has 881 bits representing element count, type of ring system, atom pairing, nearest neighbors, etc.

Hash fingerprints, such as Morgan fingerprints and functional-class fingerprints, do not have predefined substructures. Instead, they extract fragments from the given dataset and convert them into numeric values.

Due to extraction of molecular substructures and matching process, obtaining molecular fingerprints is quite complicated. Meanwhile, the representation ability of binary molecular fingerprints is limited. Though molecular fingerprints are still used in SMILES \cite{xu2017seq2seq} and molecular graphs \cite{jin2018junction}, there are less works that take molecular fingerprints as the representation approach in recent progress. 

\subsection{SMILES}

Simplified molecular input line entry system (SMILES) \cite{weininger1988smiles} is the method that uses strings to represent the molecular structure. Complex molecular configurations are transformed into sequences in vector form, which is easy to obtain. Meanwhile, with the development of natural language processing techniques, SMILES string can be regarded as a sentence for molecule representations \cite{honda2019smiles,adilov2021generative,radford2019language}. However, the SMILES representation also has certain limitations. The SMILES representation length varies with the molecule size, which poses significant challenges in devising generic models. Also, the property of sentences makes it challenging to represent three-dimensional information about molecules.

\subsection{Molecular Graph}

Graphs are realistic representations of molecules, preserving the rigorous details of the topology and geometry structures. The nodes in the graph can represent the attribute, and valence state of atoms, while the edges in the graph contain information such as the type and length of chemical bonds. Deep learning methods for graphs, especially for graph neural networks (GNNs), have promising performance in many tasks, such as property prediction and reaction prediction. 
 
Compared to SMILES and manually designed fingerprints in sequence, the graph-formed representation alleviates the order ambiguity represented in sequence form. It preserves rich structural information, such as the distances between atoms and the angles between bonds. Chen et al. \cite{chen2019utilizing} emphasize the great importance of edge features, inspiring more to be explored by preserving atom and bond information via the propagation of GNNs.

\section{Molecule Generation} 
\label{sec:Molecule Generation}

\begin{figure*}[t]
     \centering
     \includegraphics[width=0.7\linewidth]{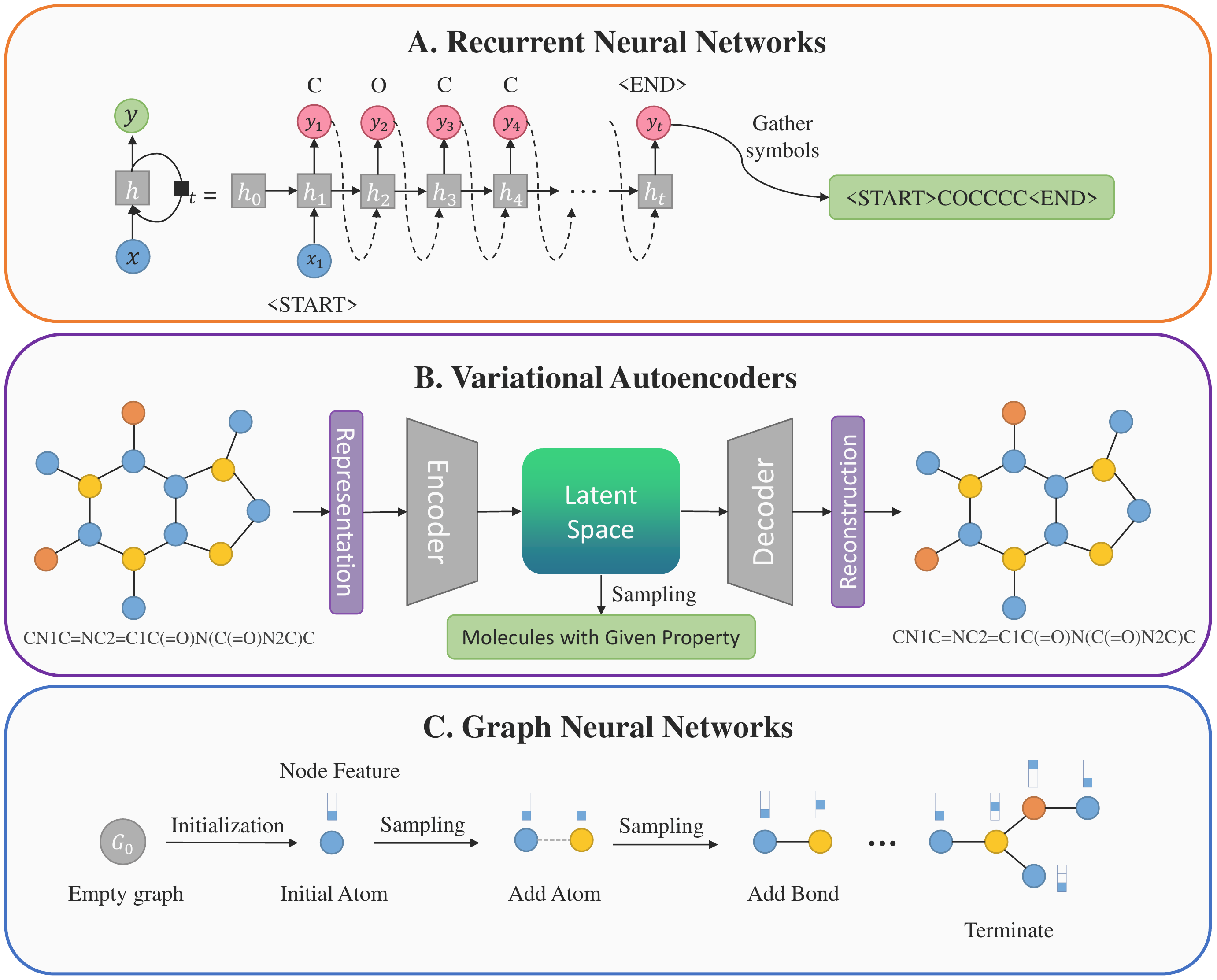}
     \caption{Illustrations of (A) Recurrent Neural Networks, (B) Variational Autoencoders and (C) Graph Neural Networks in molecule generation.}
     \label{fig:generation}
     \vspace{-10pt}
\end{figure*}

\subsection{String-based}

Based on this molecular representation method, some standard deep learning models for processing sequence information, such as recurrent neural networks (RNNs), variational auto-encoders (VAEs), and genetic algorithm, are applied to the molecular generation model. These methods have been shown to perform well for many domains in deep learning.

\textbf{Recurrent Neural Networks (RNN)}. 
RNNs generate output symbols based on input from previous steps, then these symbols are gathered to form generated molecule (Fig.~\ref{fig:generation}A).
Olivecrona et al. \cite{olivecrona2017molecular} develop a policy based RL approach to fine-tune a pre-trained RNN network.
Segler et al. \cite{segler2018generating} generate large sets of diverse molecules for virtual screening campaigns and then generate smaller, focused libraries enriched with possibly active molecules for a specific target. Stacked RNN layers are used to generate new molecular SMILES representations. This work demonstrates the robustness and transferability of RNNs in molecular generative models. 
Gupta et al. \cite{gupta2018generative} also follow the idea of pre-training on large-scale molecular datasets and fine-tuning on small-scale datasets to generate molecules of specific properties. This work also highlights the potential of RNN-based generative models to complete molecules with particular properties from molecular fragments.  
Grisoni et al. \cite{grisoni2020bidirectional} propose a method for generative and data augmentation from both directions of the sequence. Since molecular sequences are not as directional as NLP-like sequences, this attempt has been shown to perform better than unidirectional RNNs in experiments.

Many works also employ long-short term memory (LSTM) networks. Bjerrum et al. \cite{bjerrum2017molecular} train LSTM-based molecular generative models on two datasets consisting of fragment-like and drug-like molecules. 
Ertl et al. \cite{ertl2017silico} use an LSTM-based molecule generation model to generate a large number of new molecules in a short period, of which structural features and functional groups remain closed within the drug-like space defined by the bioactive molecules from ChEMBL. 

\textbf{Variational Auto-Encoders (VAE)}.
VAEs encode high-dimensional, discrete chemical space into a low-dimensional, continuous latent space, which can be used to sample molecules with given property (Fig.~\ref{fig:generation}B). Schiff et al. \cite{schiff2022augmenting} only use SMILES representations instead of graph-based molecular representations. The method is well-performed by incorporating 3D geometric information into the VAE.  
Alperstein et al.~\cite{alperstein2019all} consider the non-uniqueness of molecular SMILES expression and propose encoding multiple SMILES strings of a single molecule. VAE effectively learns the hidden information between different SMILES representations and performs well in molecule generation and optimization in fully supervised and semi-supervised situations.

In addition to the commonly used RNN and VAE models, Nigam et al. \cite{nigam2019augmenting} combine the traditional genetic algorithm with deep neural network. The DNN-based discriminator improves the diversity of generated molecules and increases the interpretability
of genetic algorithm. Moss et al. \cite{moss2020boss} first use string kernels and genetic algorithms in bayesian optimization loops. Most bayesian optimization approaches need to convert the original string inputs to fixed-size vectors, while this architecture can directly work on raw strings.

In conclusion, string-based molecule generation considers the input and target molecules as the sequence data. Deep learning models like RNN, VAE, and genetic algorithm achieve great success in the generation task. However, there is still a limitation for the string-based method since it may lose the complex geometry information of molecules. Thus, most current works focus on the graph-based model.

\subsection{Graph-based}
Motif, also referred to as scaffold or rationale, is defined as the molecule fragment. Motif corresponds to the concept of functional groups in chemistry, closely related to the molecule's chemical properties. Thus, motif can provide more prior knowledge when generating molecules with given properties. Meanwhile, motif already has a specific valid structure. With the involvement of motif, generating invalid molecules can be avoided to some extent. Motif-based molecular generation treats molecules as a collection of fragments. Substructures are then spliced to generate new molecules. Non-motif molecular generation generates molecules atom by atom (Fig.~\ref{fig:generation}C).

\textbf{Motif-based Molecular Generation}.
Jin et al. \cite{jin2018junction} generate molecular graphs using a variational auto-encoder. They first extract valid chemical substructures from the training set to generate a tree-structured scaffold, then use the subgraphs as building blocks and combine them to form a molecular graph. Compared with the node-by-node approach, this structure-by-structure approach maintains chemical validity at each step. 
Jin et al. \cite{jin2018learning} further combine variational auto-encoder with an adversarial training method to obtain various and valid output distributions. An adversarial regularization is included to align the distribution of generated graphs with the distribution of valid targets.
Jin et al. \cite{jin2020hierarchical} propose a hierarchical graph encoder-decoder that can generate large molecules such as polymers. Each molecule has a fine-to-coarse representation from atom level to
motif level.
Similar to \cite{jin2018junction}, Maziarz et al. \cite{maziarz2021learning} use extracted motif to generate molecules. Meanwhile, they integrate this method with atom-by-atom generation.
You et al. \cite{you2018graph} develop a graph convolutional network through reinforcement learning. The generation procedure is regarded as Markov Decision Process, and domain-specific rewards are incorporated into the model.
Yang et al. \cite{yang2021hit} also propose a reinforcement learning framework that generates pharmacochemically acceptable molecules with significant docking scores. This method chooses a chemically realistic and pharmacochemically acceptable fragment based on the given state of the molecule. 
To generate molecules with multiple property constraints, Jin et al. \cite{jin2020multi} extract rationales for each property and combine them as multi-property rationales. A graph completion model is then implemented to generate complete molecules from rationales.
Based on \cite{jin2020multi}, Chen et al. \cite{chen2021molecule} propose an Expectation-Maximization like algorithm. In the E-step, an explainable model extracts rationales from candidate molecules. While in the M-step, rationales are completed to generate molecules with a higher property score. 
Xie et al.~\cite{xie2021mars} employ the annealed Markov chain Monte Carlo sampling method \cite{andrieu2003introduction} to search the vase chemical space. They further train a graph neural network based on the sampling result to choose proper candidate edits. 
Fu et al. \cite{fu2021differentiable} propose differentiable scaffolding tree which is a gradient-based optimization method.  Graph convolutional network transforms the discrete chemical space into a locally differentiable space. Further, the molecule is optimized by gradient back-propagation from the target.

\textbf{Non-motif Molecular Generation}.
Zhou et al. \cite{zhou2019optimization} generate molecule atom by atom with the help of the double Q-learning technique. This method does not require pre-training, so that the training set does not limit exploration capability.
Combining autoregressive and flow-based methods, Shi et al.~\cite{shi2020graphaf} develop a flow-based autoregressive graph generation model, allowing parallel computation in the training process. The model generates edges and nodes concerning the current graph. In each generation step, chemical knowledge like the valence rule could be utilized to check the molecular validity. 
For molecule optimization problem, Korovina et al.~\cite{korovina2020chembo} develop a bayesian optimization model that focuses on small organic molecules. Chen et al. \cite{chen2021cells} propose a cost-effective evolution strategy in latent space. An evolutionary algorithm is introduced into the latent space to search for the desired molecules.
Wu et al. \cite{wu2021spatial} propose distilled graph attention policy network, which is aimed to optimize molecule structure based on user-defined objectives. A spatial graph attention mechanism is introduced, considering self-attention over the node and edge attributes.

In addition to exploring 2D molecule generation, many works have been done on 3D molecule generation tasks. Simm et al. \cite{simm2019generative} combine a conditional variational autoencoder with a euclidean distance geometry algorithm. This method uses a set of pairwise distances between atoms to describe molecule conformation instead of cartesian coordinates. Following \cite{simm2019generative}, Xu et al. \cite{xu2021end} first propose an end-to-end conditional variational autoencoder framework for molecular conformation prediction. In this work, the distance prediction problem and the distance geometry problem are simultaneously optimized via bilevel programming.
Shi et al. \cite{shi2021learning} directly estimate the gradient fields of the log density of atomic coordinates. 
The estimated gradient fields can generate conformations directly via Langevin dynamics, thus reducing the computation error. 
Zhu et al. \cite{zhu2022direct} design a loss function invariant to the roto-translation of coordinates of conformations and permutation of symmetric atoms in molecules.  
Xu et al. \cite{xu2022geodiff} analogize conformation generation to diffusion process in thermodynamics. Atoms are first stabilized in specific conformations and then diffuse into a noise distribution. The reverse process is regarded as Markov chain solving for conformation generation problems.
Luo et al. \cite{luo2021autoregressive} propose an autoregressive flow model which can generate 3D molecules. This method directly generates distances, angles, and torsion angles of atoms rather than 3D coordinates. Once a molecule is generated, the relative distances and angles between atoms are determined, thus ensuring invariance and equivariance.
Mansimov et al. \cite{mansimov2019molecular} propose a conditional deep generative graph neural network to learn the energy function of molecule conformations. Compared with conventional molecular force field methods, this model can directly generate energy-supported molecular conformations without requiring multiple iterations.
Guan et al. \cite{guan2021energy} also propose an energy-inspired neural optimization formulation. The molecule conformation is optimized by gradient descent for 3D coordinates through the energy surface.
Xu et al. \cite{xu2021learning} propose a method that combines flow-based and energy-based models. Flow-based model is used to generate a distance matrix from a Gaussian prior. Then the energy-based model searches and optimizes the 3D coordinates of the molecule.
\section{Molecular Property Prediction}
\label{sec:Molecular Property Prediction}

\begin{figure*}[t]
     \centering
     \includegraphics[width=0.7\linewidth]{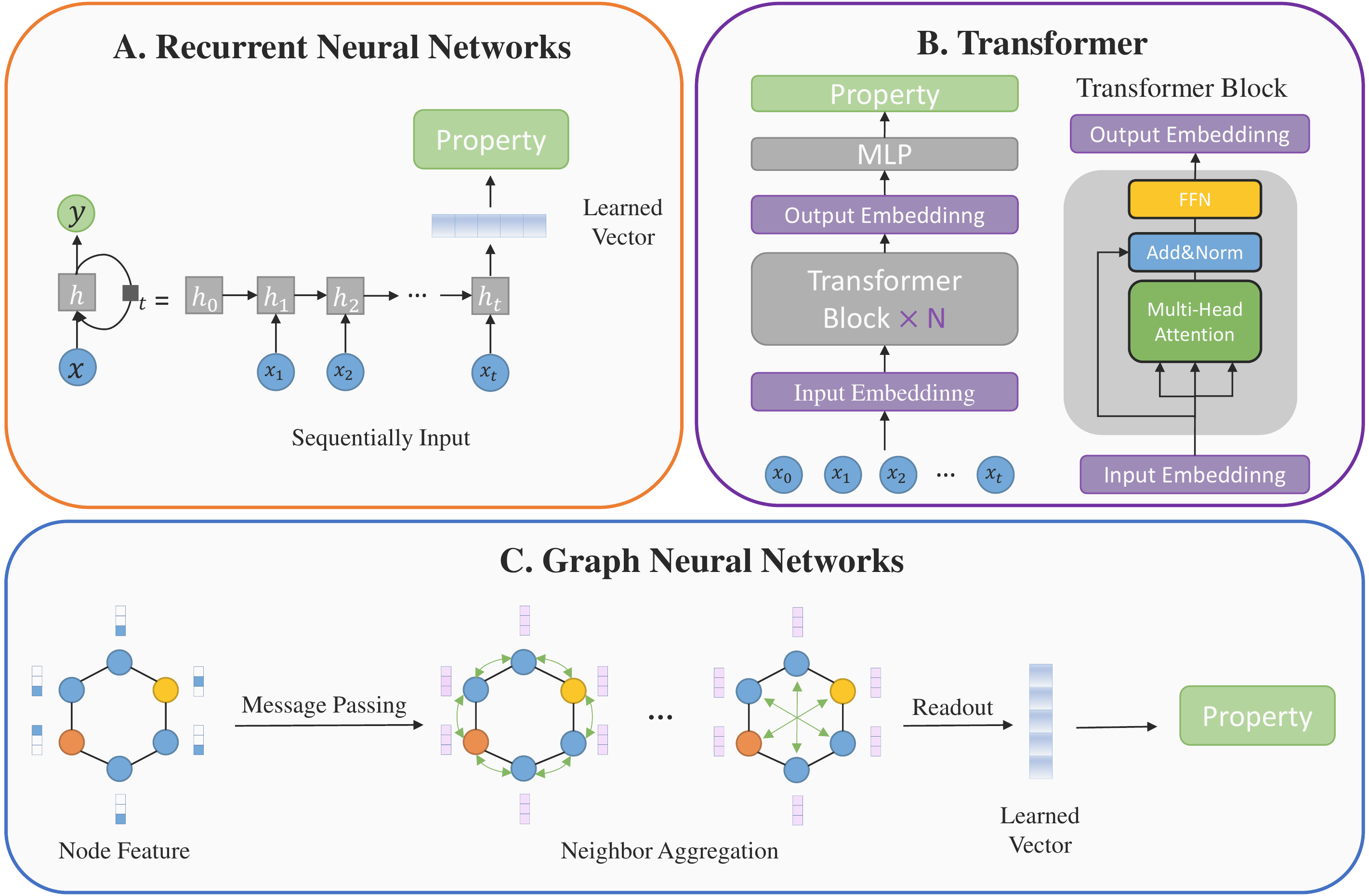}
     \caption{Illustrations of (A) Recurrent Neural Networks, (B) Transformer and (C) Graph Neural Networks in molecular property prediction.}
     \label{fig:prediction}
     \vspace{-10pt}
\end{figure*}

\subsection{SMILES-based}
The SMILES-based methods utilize various architectures, such as RNN, VAE, and transformers, which are demonstrated as follows.

\textbf{Recurrent Neural Networks (RNN)}.
RNNs are shown to be effective designs for sequential input such as text data. Thus, RNN and its two variants, LSTMs \cite{hochreiter1997long} and GRUs \cite{cho2014properties}, are applied to learn chemical properties from SMILES (Fig.~\ref{fig:prediction}A).

Mayr et al.~\cite{mayr2018large} employe long short-term memory networks to construct the SmilesLSTM architecture, which utilizes SMILES strings as inputs.
Inspired by the success of RNNs in sequence-to-sequence language translation, Goh et al. propose a sequence-to-vector model SMILES2vec \cite{goh2017smiles2vec}, where the sequence is SMILES, and the vector is measured property. Instead of explicitly encoding the molecular information from element compositions in SMILES, they explore four CNN-based and RNNs-based architectures to learn the desired features directly. Meanwhile, an explanation mask is developed to localize the most important characters, improving the model's interpretability. 

While Goh et al. \cite{goh2017smiles2vec} stick to the commonly-used canonical SMILES, which preserves a one-to-one mapping from molecule to string, Li et al. \cite{li2022novel} make use of multiple SMILES strings of a single molecule as data augmentations, which helps to learn better grammatical features. The proposed model generates the molecular representation from multiple SMILES sequences that first convert to a one-hot vector and then follow the message-passing process in the stacked CNN and RNN architecture.

Considering that chemical information is able to canonicalize SMILES strings, Peng et al. \cite{peng2019top} exploit a framework called TOP, combining the usage of SMILES and predefined properties, including root atom positions and isomeric features. A bidirectional gated recurrent unit-based RNN (BiGRU) is employed for the SMILES strings to capture local and global context information. At the same time, a fully connected neural network (FCN) is used to process the physicochemical feature vector. When performing on balanced datasets, TOP significantly outperforms other methods, including SMILES2vec~\cite{goh2017smiles2vec}, CheMixNet \cite{paul2018chemixnet}, and Chemception \cite{goh2017chemception}, which uses only the images of 2D molecular drawings decoded from SMILES.

\textbf{Variational Auto-Encoder (VAE)}.
VAE models share a similar structure to the encoder-decoder structure mentioned in the RNN-based architecture. However, it is different in formulating the assumption that the embedded space follows Gaussian distributions. Among the existing architectures, VAEs focus more on molecular-based latent representations instead of specifying prediction tasks. Adapting the SSVAE devised in \cite{kingma2014semi}, Kang and Cho feed SMILES as the input variables and its continuous-valued property vectors as outputs. The encoder and decoder networks perform the recovery task of SMILES with the learned latent molecular representation as an intermediate between the two stages. Once the SSVAE is trained, the remaining RNN acts as the predictor network. ALL SMILES VAE \cite{alperstein2019all} ensures the learned latent molecular representation to capture the molecular feature instead of its SMILES realization by using multiple SMILES and implicitly performing message passing among spanning trees of the molecular graph via recurrent structures. 

\textbf{Transformer}.
Transformer architecture follows a similar encoder-decoder process, but it does not employ recurrent connections like RNNs and is thus prone to be more stable to converge and superior to RNNs in featurization on longer and larger corpus (Fig.~\ref{fig:prediction}B) \cite{honda2019smiles,adilov2021generative}. To guarantee the learned molecular representations are able to successfully transfer in numerous downstream tasks, pre-training strategies on a large SMILES corpus stand out. SMILES-GPT \cite{ adilov2021generative}, prolonging the pre-training approach, applies a lightweight adapter network after each attention block on the design of GPT-2 \cite{radford2019language} to learn specific task patterns. In this way, it alleviates the loss of domain knowledge when transferring the pre-trained model to the separate downstream task.

The language representation model BERT \cite{devlin2018bert} introduces masked language model (MLM) pre-training techniques that alleviate the constraints of incorporating context from uni-direction. Benefiting from the similarities between MLM and atom masking in molecules, the BERT-style architecture has emerged as a robust technique in molecular representation learning. The efficacy of BERT-liked architectures, such as RoBERTa \cite{liu2019roberta}, is proved to achieve comparatively high classification accuracy when training through an extensive database \cite{rahimovich2021predicting}. With a careful exploration of pre-training dataset size, tokenizer, and string representation, ChemBERTa \cite{chithrananda2020chemberta} adapted from RoBERTa \cite{liu2019roberta} again emphasizes the efficiency of large-scale pre-training in molecular property prediction.

\subsection{Fingerprint-based}
Fingerprints, as another class of molecular representations, also benefits molecular predictive tasks.
Unterthiner et al.~\cite{unterthiner2014deep} employ multi-task learning in molecular property prediction, leveraging the power of deep neural networks to generate hierarchical representations of compounds. 
Meanwhile, Mayr et al.~\cite{mayr2016deeptox} introduce the DeepTox pipeline. DeepTox employs normalization techniques to generate standardized chemical representations of compounds, and subsequently computes an extensive array of chemical descriptors that serve as input to machine learning algorithms.
Traditionally hand-crafted fingerprints are also introduced as the complement of SMILES strings for molecular feature information. In this way, the model performance no longer fluctuates with the length of SMILES strings drastically. Paul et al. \cite{paul2018chemixnet} harness two successful architectures for property prediction, including SMILES2vec and MLP, into a multi-input-single-output (MISO) architecture CheMixNet. The model takes both SMILES strings and MACCS fingerprints as inputs. 
Leveraging both types of molecular structural representations as parallel inputs increase generalizability. 
Schimunek  et al.~\cite{schimunek2022context} present a novel few-shot method that enhances the representation of a molecule by incorporating information from known context or reference molecules.

However, the performance of traditional fingerprints is not good enough in prediction tasks due to the restrictions of the manually designed methods. For example, the hash-based fingerprint brings limited task-specific information due to its irreversibility. With the advancement of deep-learning techniques, novel methods have been invented to generate deep-learning-based fingerprints for property prediction.

\textbf{Recurrent Neural Networks (RNN)}.
RNNs-based unsupervised fingerprint methods are proposed to tackle the problem of scarce labeled data. Deep neural networks are applied to utilize a large pool of unlabeled data. 
Motivated by the breakthrough success of the model applied in seq-to-seq language translations, encoder-decoder structured neural networks have been popular in generating fingerprints based on sequence-like SMILES strings. Xu et al. \cite{xu2017seq2seq} extract the intermediate fixed-size molecular feature vectors as seq2seq fingerprints during the process of mapping a SMILES string to the desired vector and then translating back to the original string. The extracted fingerprints, combined with SMILES labels, are trained for prediction tasks in supervised methods with classifiers or regressors, such as multi-layer perceptron. Based on the design, Zhang et al. \cite{zhang2018seq3seq} modify it to the first end-to-end framework coupling recovery and inference tasks in a model named seq3seq fingerprint.

\textbf{Transformer}.
Inspired by the excellent performance of the self-attention mechanism in language tasks, Transformer is believed to have a good performance on abundant unlabeled molecular data. By adopting the pre-training approach that shows promising results in the NLP field, SMILES transformer~\cite{honda2019smiles} is presented to learn molecular fingerprints through large-scaled pre-training on canonical SMILES. The data-driven extracted fingerprints work well with a simple multi-layer perceptron, which shows the great potential of exploiting unsupervised pre-training in property prediction.

BERT, similar to the encoder of the Transformer, is also a promising structure. SMILES-BERT \cite{wang2019smiles}, following the BERT approach, randomly masks selected tokens in an input SMILE to the mask token, any other token in the dictionary, or unchanged according to an 85/10/5 split ratio. The fingerprints generated from the MLM strategy are experimentally proved to have higher predictive accuracy on a rather large dataset than some representative fingerprints, including circular fingerprint \cite{ glen2006circular}, neural fingerprint \cite{bahdanau2014neural} and seq2seq fingerprint \cite{xu2017seq2seq}.

\subsection{Graph-based}

\textbf{Graph Convolutional Network (GCN)}.
Graph convolutional networks (GCN) \cite{kipf2016semi} are always considered a common baseline choice for applications. The model focuses on learning node states and aggregates the neighboring messages (Fig.~\ref{fig:prediction}C). A wide range of applications in molecule representation learning is based on GCNs due to its lightweight calculation and scalability to large graphs. 

Labeled datasets, for instance, the QM9 dataset \cite{ramakrishnan2014quantum} is composed of molecular properties approximated by costly methods such as density functional theory. The neighborhood aggregation scheme of GCNs is thus frequently used to exploit an overall high-order representation learning for molecules. A multilevel graph convolutional neural network proposed in \cite{lu2019molecular} learns the node representations by preserving the conformation information with hierarchical modeling (atom-wise, pair-wise, etc), and its spatial information by introducing a radial basis function layer for robust distance tensors. The interaction layers that effectively combine self features with collected messages from neighbors are applied for high-order atom representations.

It is often challenging that annotated labels are so expensive that datasets are commonly seen to be partially labeled in real-world applications. Training on such limited labeled data easily leads to over-fitting and poor performance of dissimilar molecules from the training data. To address the problem of scarce task-specific labels and out-of-distribution predictions, researchers have made great efforts in pre-training to leverage the unlabeled data and improve the generalization power of GNNs. The key to pre-training is finding a suitable and effective task to leverage many unlabeled structures \cite{liu2021pre}, yet choosing effective pre-training strategies is challenging since the selected properties must be aligned with the interests of specific downstream tasks. Otherwise, as mentioned above, the ``negative transfer'' effect can seriously harm the generalization when transferring the pre-learned knowledge to a new downstream task \cite{hu2019strategies}.  

Pre-training is a common and effective strategy for CNNs,  ImageMol\cite{zeng2022accurate}
though the robust scheme has not been explored for GNNs until recent years. Hu et al. \cite{hu2019strategies} conduct the first systematic large-scale investigation of pre-training strategies for GNNs. They propose self-supervised context prediction and attribute masking methods at the node level and supervised task predictions for domain-specific information decoding. Along with the foremost introduction of pre-training strategies for property prediction models, Hu et al. \cite{hu2019strategies} apply context prediction and attribute masking with various GNN structures pre-trained on eight datasets from MoluculeNet, proving the efficacy of node-level strategies,. These methods open new ways for GNN pre-training schemes yet are far from satisfactory. Rong et al. \cite{rong2020self} argue that the graph-level tasks are impeded with limited labels and introduce more risks of negative transfer in downstream tasks. Also, isolating the context and node type predictions as labels makes it hard to preserve local structure knowledge or address highly frequent atoms.  
To achieve better generalizability, some have made great efforts to investigate pre-training with data augmentation. The proposed model GraphCL \cite{you2020graph} performs pre-training through maximizing MI between two augmented representations in the latent space generated by different data augmentation methods.

Inspired by these generalized techniques, Wang et al. \cite{wang2022molecular} introduce the specific molecular-designed model MolCLR that addresses GNNs pre-training with graph augmentations by contrastive learning. It follows a similar manner as GraphCL \cite{you2020graph} by maximizing distance between two latent augmented vectors resulting from two stochastic augmentations. However, augmentation methods are more carefully designed and more specified for molecules. Despite atom masking and subgraph removal, the model utilizes bond deletions instead of random perturbations that can mimic the reactions of chemical bonds to learn more valuable features. However, researchers are still doubting the effectiveness of data augmentation in pre-training. Li et al. \cite{li2021geomgcl} argue that random modification of atoms and edges can harmfully destroy the natural structure. 
Liu et al.~\cite{liu2019n} propose the N-gram graph, which is a unsupervised representation for molecules. The approach commences by embedding the vertices in the molecule graph. It then creates a concise representation for the graph by assembling the vertex embeddings in short walks within the graph. Altae et al.~\cite{altae2017low} develop a architecture that combines iterative refinement long short-term memory with graph convolutional neural networks. This one-shot learning approach allows for significant reductions in the amount of data needed to make accurate predictions.

\textbf{Messsage Passing Neural Network (MPNN)}.
When inferring molecular properties, the impact of bond types and distances of atoms can not be neglected. GCNs, though considering neighborhood aggregation, are not able to distinguish different edge types. To better handle edge features, Gilmer et al. \cite{gilmer2017neural} formally propose a message passing neural network (MPNN) to learn a graph level embedding with node features as well as the weighted edge messages. The framework involves two phases, a message-passing phase and a readout phase. By modeling atoms as nodes and bonds as edges, feature vectors of nodes and edges are input to the model as initialization. Further, message are aggregated from neighboring nodes through multiple layers of message passing. The readout phase then summarizes the feature vectors of the whole graph to target the downstream molecular property prediction task. Variations based on MPNN \cite{gilmer2017neural} have made great progress, such as message passing with attention mechanism in \cite{rong2020self}, spherical message passing for 3D molecular graphs \cite{liu2021spherical} and so on. Despite the focus on node states, preservation of important bond (edge) information is also taken into account for better performance \cite{chen2019utilizing}.

The message passing scheme has shown its superior performance in molecular property prediction. Hao et al. \cite{hao2020asgn} make the first attempt to employ the powerful framework for molecular property prediction in a semi-supervised manner. The novel strategy, by finding the most diversified subset in the unlabeled set and continuously adding them to the labeled dataset for training, alleviates the over-fitting and imbalance between labeled and unlabeled molecules.

Under the assumption of message passing neural networks, two atoms in the same neighborhood should have similar representations, neglecting that their positions in the molecule are distinct. Therefore, MPNNs are usually armed with positional encoding to alleviate the limitations. Instead of applying higher-order representations, Dwivedi et al. \cite{dwivedi2021graph} introduce positional encoding (PE) for nodes and leverage the information as input information to learn both positional and structural feature representations at the same time. Combining PE with MPNNs contributes to more robust node embedding since it fills the gap in the canonical positioning of nodes. 

While 2D molecule graph provides rich information on topology, 3D geometric views also play a vital role in predicting molecular characteristics. Based upon the original message passing network \cite{gilmer2017neural}, Liu et al. \cite{liu2021spherical} propose the spherical message passing and SphereNet for 3D molecular learning. The model performs message passing in the spherical coordinate system and uses the relative 3D information and torsion computation to generalize invariant predictions of rotation and translation. Furthermore, it is shown that 3D geometric views significantly contribute to robust representation learning. In GeomGCL \cite{li2021geomgcl}, the efficiency of 3D geometry is illustrated to be superior to other methods that confine to topology structures. Li et al. \cite{li2021geomgcl} devise a method to leverage 2D and 3D pair information to improve generalization ability. The dual-channel geometric message passing architecture allows collaborative supervision of 2D and 3D views between each other. To further leverage the local spatial correlations, a regularizer on angle domains is applied for generalization on similar property prediction. These strategies help achieve the goal of capturing both chemical semantic information and geometric information.

\textbf{Graph Isomorphism Network (GIN)}.
Capturing statistical dependencies is essential in molecular presentation, and the contrastive method is among the most effective approaches to obtaining the features \cite{velickovic2019deep}. Borrowed the idea of mutual information neural estimator (MINE) from mutual information (MI), deep InfoMax (DIM) \cite{hjelm2018learning} is introduced to maximize information content between input and output. Inspired by DIM, deep graph InfoMax (DGI) \cite{velickovic2019deep} trains a contrastive node encoder that uses GCN to maximize local mutual information to capture global information. However, the GCN convolutional encoding fails to distinguish some graph structures. Instead of mean and max aggregation, a more careful design of neighbor aggregation is needed for more substantial discriminative power for graph representations.

Graph isomorphism network (GIN) \cite{xu2018powerful} is then proposed and shown to be as powerful as Weisfeiler-Lehman graph isomorphism test that determines whether two graphs are topologically identical. The structure using sum aggregation has better performance in discriminating graph instances. Sun et al. \cite{sun2019infograph} replace the graph convolution encoders with GIN and expands the mutual information idea in semi-supervised learning scheme using a ''student-teacher'' model to train the whole graph level embedding. 

GINs are frequently used for backbone models in recent 2D GNN designs for molecular applications. Liu et al. \cite{liu2021pre} consider 3D geometric views for pre-training with message passing, aiming to train the 2D molecular GIN encoder to recover its 3D counterparts to some extent. By maximizing MI between 2D topological structure and 3D spatial and geometric information, the pre-trained model can benefit from implicit 3D geometric prior information, even if no 3D structure is provided in the downstream task. Meanwhile, a comparisons on two pre-training models of GraphCL \cite{you2020graph} and GraphMVP are also illustrated in \cite{liu2021pre}, showing that incorporate both 2D topology and 3D geometry is more competitive.

\textbf{Recurrent Graph Neural Networks}. 
Recurrent neural networks, such as RNNs, GRUs, and LSTMs, are powerful tools for processing vary-sized sequences. Converting graphs to sequences is a potential solution to get rid of the fixed-size matrix problem for molecule graphs. Popular recurrent units, GRUs and LSTMs, have been widely used in molecule generation. Furthermore, recurrent units play a significant role in message aggregation and conformation preserving.

The gated recurrent units are responsible for updating hidden representations of nodes in message passing. While the edge network aggregates messages from 1-hop neighbors of the nodes, the GRU updates the node states with the most recent node state resulting from aggregated message based on the previous states \cite{gilmer2017neural}, therefore, allowing information to travel through connected bonds. The architecture also allows preserving information from long ago. Li et al. \cite{li2020conformation} propose a conformation-aware architecture with GRUs, namely HamNet. With modifications in message calculations, the HamNet incorporates relative positions and momentum in atom representations generated from GRU.

The message passing scheme of the GCNs follows that adjacent atoms are in an identical chemical environment, \textit{i.e.}, neighborhood should have similar representation. Although the deep neural network can capture the local and global structures, it may fail to distinguish atoms performing different roles yet having the same neighborhoods. Therefore, the topological order usually needs to be introduced to the design by leveraging positional encoding techniques \cite{dwivedi2021graph}, or ordered SMILES representation. With an LSTM conducted over the GCN outputs, the unique positions for atoms can be determined by controlling the order of atoms in the LSTM that coincides with the canonical SMILES \cite{li2020conformation,yang2021deep}.

\textbf{Transformer}.
Multiple prior works have applied transformer-style architecture in molecular applications. The expressive power of transformer trained on sequence representations, SMILES, and fingerprints has significantly been explored \cite{honda2019smiles,liu2019roberta,chithrananda2020chemberta}, and the architecture is also adapted to molecular graph structures to enhance the representational power. Maziarka et al. \cite{maziarka2020molecule} introduce a novel pre-trained molecule attention transformer that learns domain-knowledge from the pre-text task and then utilizes the knowledge to augment self-attention. The key innovation of the proposed model is the replacement of the molecular multi-head self-attention layers that integrate the molecule graph and inter-atomic distances with the self-attention. Both sources of information extracted from graph structures further improve the performance. Rong et al. \cite{rong2020self} follow a similar pre-train idea and pay more attention to bi-level information extraction from molecule graphs to better preserve the domain knowledge and avoid a negative transfer. Dynamic message passing networks are applied to extract local subgraph information from node embedding, while transformer modules are introduced respectively for nodes and edges to extract global relations of nodes. Transformer-style architecture with a dynamic message passing mechanism has an excellent performance in capturing structural information and enhances the expressive power of GNN. Future work may explore better pre-training tasks for the transformer-style architecture to help with various downstream molecular tasks.

\section{Retrosynthesis}
\label{sec:Retrosynthesis}

\begin{figure*}[t]
     \centering
     \includegraphics[width=0.68\linewidth]{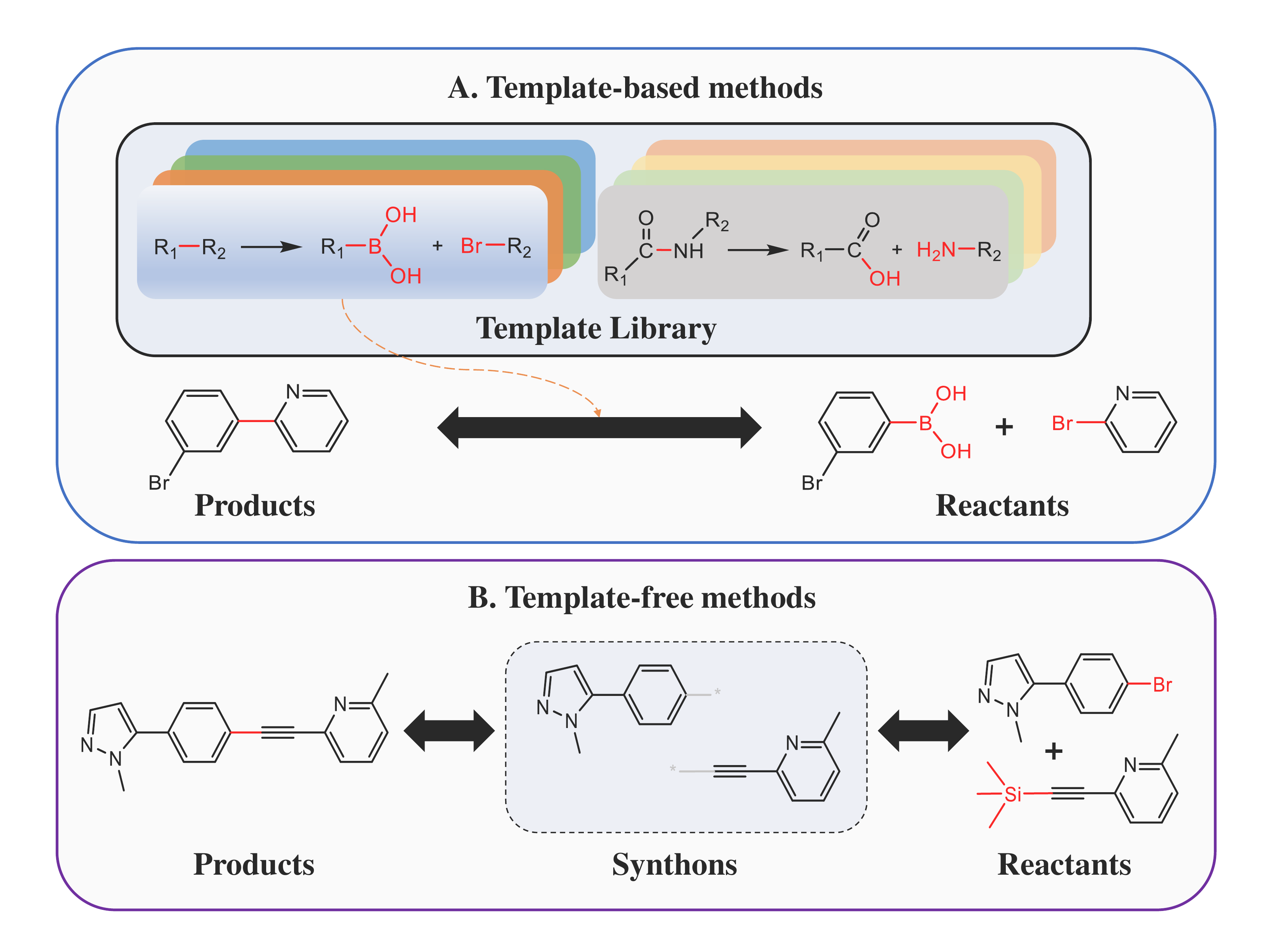}
     \caption{Illustrations of (A) Template-based and (B) Template-free Methods in Retrosynthesis and Reaction Prediction.}
     \label{fig:retro}
     \vspace{-10pt}
\end{figure*}

The goal of retrosynthesis is to decompose a complicated target product to find a set of reactant molecules with relatively simple structures. It is a fundamental task in synthesis planning and drug manufacturing. For example, a common prescription drug—aspirin, can be chopped up into two synthons, easily obtained from a nucleophilic addition-elimination reaction. However, the enormous potential searching space for all possible combinations results in the high cost of computational complexity. With the significant development of artificial intelligence, scientists have devoted themselves to finding efficient and effective computer-aid methods in retrosynthesis analysis.  

This section analyzes works that apply machine learning methods to solve the retrosynthesis prediction. We will introduce some early methods based on reaction templates and then analyze the template-free models for each task. 


\subsection{Template-based}

Template-based methods rely on encoding reaction templates or rules, which are generated either manually or automatically from the currently existing databases. Those templates are used to derive the retrosynthetic precursors. As there are different possible reaction templates for one target molecule, people developed various algorithms which consider chemical context, possible trade-offs, and reactivity to decide the most appropriate rule (Fig.~\ref{fig:retro}A). 

\textbf{String-based}.
String-based methods convert molecules to fingerprints or SMILES and then put them into the network. Segler and Waller \cite{segler2017neural} propose a neural symbolic model which uses a deep neural network to predict the possible reaction template. As the encoding reaction templates are finite, they view retrosynthesis analysis as a multi-class classification task. They represent the input target molecule as Morgan fingerprint, and the output transformation will be applied to the target molecule to get the retrosynthetic precursors. Because similar molecules may have similar reactions, Coley et al. \cite{coley2017computer} introduce a similarity-based approach. This method uses similarity as the metric to determine the precedent reactions, which are used to find the appropriate reaction site for the target molecule. It then applies the corresponding templates to yield candidate precursors, which convert to reaction molecules by further applying similarity calculation. Baylon et al. \cite{baylon2019enhancing} present a multi-scale approach with a deep highway network, which groups reaction rules based on chemical similarities without supervision. The model determines the reaction rule in two stages. It first decides which group the target molecule belongs to and then predicts the specified transformation rules. Segler et al. \cite{segler2018planning} view retrosynthesis as a Markov decision process. They combine Monte Carlo tree search with different neural networks, which guide the search and select the proper retrosynthesis direction. Seidl et al.~\cite{seidl2022improving} present a novel approach to single-step retrosynthesis modeling using Modern Hopfield Networks. Their template-based model utilizes encoded representations of both molecules and reaction templates to accurately predict the relevance of templates for a given molecule.

\textbf{Graph-based}.
Graphical models utilize the expressiveness and scalability of neural networks to extract useful information. Dai et al. \cite{dai2019retrosynthesis} apply conditional graph logic network, which views chemical templates as logical rules. In logical rules, graphical structures of molecules and subgraphs are considered as logical variables. The predicted outcome is then formulated as the joint probability of the reactants and the rules. Instead of predicting reactants globally, Chen et al. \cite{chen2021deep} propose LocalRetro, which focuses on local reaction templates. Each atom and bond uses a message passing neural network to learn the local reactivity and applies several particular layers to learn the remote chemical knowledge. Based on the predicted atoms and bonds, local reaction templates are then applied to get the final reactants.

\subsection{Template-free}

As deep learning has developed significantly in recent years, some template-free methods based on deep learning strategies have become increasingly popular. Template-free methods transform chemical structures from one to others without any reaction rules (Fig.~\ref{fig:retro}B). 

\textbf{String-based}.
As machine translation and retrosynthesis transform the source to target languages or molecules, researchers propose various models based on such an analogy. Those models utilize SMILES notation to represent molecules as strings. Liu et al. \cite{liu2017retrosynthetic} build a LSTM-based sequence-to-sequence model, where the inputs are the target molecule and the specific reaction type, and the output is the most probable corresponding reactants. The advent of the transformer offers exciting opportunities for studying its various applications. Karpov et al. \cite{karpov2019transformer} utilize the transformer model and applies weights averaging and snapshot learning to complete the task. In order to make the prediction more general and diverse, Chen et al. \cite{chen2019learning} propose several pre-training approaches which generate new examples by randomly disconnecting the bonds or following the templates. They also use a mixture model with many latent variables to increase the diversity of the possible outcome reactants. Ishiguro et al. \cite{ishiguro2020data} further improve seq2seq models by applying the idea of data transfer to retrosynthesis. They consider different methods, including joint training, self-training, pre-training, and fine-tuning, and evaluate their performance using augmented datasets. Zheng et al. \cite{zheng2019predicting} develop a self-corrected retrosynthesis predictor which applies the reaction predictor to get raw candidate reactants. It uses another transformer that mimics the grammar corrector to fix molecular syntax errors.

Zhao et al. \cite{zhao2022leveraging} utilize reaction-aware substructures--which will not change during the chemical interactions to predict retrosynthesis transformation. Similar to previous works, they utilize a dual transformer encoder structure to get a product to candidate reactants mapping. They further isolate the unchanged substructure from those candidate reactants based on their fingerprints and use a sequence-to-sequence model to predict the output.

\textbf{Graph-based}.
Besides the popularity of sequence models, various graphical models are proposed with their unique advantages in the expressiveness of subgraph structures and the rapid development of graph neural networks. Sacha et al. \cite{sacha2021molecule} present Molecule Edit Graph Attention Network, which views the retrosynthesis as a sequence of the graph editing process. The encoder-decoder structure model outputs a set of graph actions, including editing and adding atoms, bonds, or rings. Those actions are then applied to the target molecule to generate the desired reactants. Instead of performing retrosynthesis as a single-step task, G2Gs, proposed by Shi et al. \cite{shi2020graph}, dissembles it into two parts. Their model first finds the reaction center by ranking the reactivity scores of atom pairs and then chops up the input target molecular graph into several sub-graphs. By applying a series of variation graph translations with a latent space for increasing diversity, it generates reactant graphs from those synthons. GraphRetro, built by Somnath et al. \cite{somnath2021learning}, also utilizes the two-step framework. Instead of scoring the atom pairs, GraphRetro ranks both bonds and atoms by a message passing network to predict the possible synthons. Unlike G2Gs that directly translates graphs to graphs, this model views the process of producing reactants from synthons as a classification problem and selects leaving groups from precomputed semi-templates to complete synthons. Another model, Semi-Retro \cite{gao2022semiretro}, further adds local structures of synthons into semi-templates and proposes a directed relational graph attention layer for identifying the reaction center. 

\textbf{Combination of Graph and String}.
In order to incorporate the knowledge from both graph representation and sequence representation, some researchers propose methods that balance both properties. Like G2Gs, Yan et al. \cite{yan2020retroxpert} propose RetroXpert, which splits retrosynthesis into two stages. RetroXpert uses Edge-enhanced Graph Attention Network to find the reaction centers among bonds, and it has an auxiliary task to predict the number of disconnects. In the second stage, it converts synthon graph into SMILE notation to apply the sequence model to generate the final reactants. Sun et al. \cite{sun2020energy} propose an energy-based framework for both graphical and sequence models by designing different energy functions. Based on the duality of forwarding reaction and retrosynthesis, they also develop a dual EBM variant model. By researching the intrinsic connection between graph neural network and transformer, Graph Truncated Attention, proposed by Seo et al. \cite{seo2021gta}, adds molecular graph information to the attention layers of the transformer, which combines two representations smoothly without introducing additional parameters. Alternatively, Retroformer, presented by Wan et al. \cite{wan2022retroformer}, uses a local attention head, which integrates graphical and sequence information, to learn local reactivity and global context simultaneously.
\section{Reaction Prediction} 
\label{sec:Reaction Prediction}

While the inverse processes of reaction generate massive convenience for the chemical industry, the forward synthesis of the organic reaction also raises tremendous interest. As another fundamental task in synthesis planning, reaction prediction is a task to predict the possible compounds for a given set of reactants, reagents, and solvents. Reaction prediction not only assists the evaluation for retrosynthesis but also helps with expensive and time-consuming experiments, especially those potentially harmful to laboratory technicians.

\subsection{Template-based} 

Like retrosynthesis analysis, template-based models for reaction prediction are built on reaction templates. As the search spaces for the given reactants to generate compounds are pretty big, with substantial computational costs, reaction templates can solve this problem by restricting the space. Prediction models use different ways to match the appropriate templates and various procedures to decide the outcome molecules.
 
\textbf{String-based}.
Some template-based models represent molecules as strings and then use string-based models to predict results. Using fingerprints representation of reactants and reagents as input, Wei et al. \cite{wei2016neural} present a model that first explores the application of machine learning in this task. The model predicts the possible reaction type as output, but only 16 kinds of reactions are considered in their work. By obtaining larger sets of reaction templates, more works are proposed. 
As described in the previous section, the model presented by Segler and Waller \cite{segler2017neural} can apply to both retrosynthesis analysis and reaction prediction. Specifically, their model utilizes neural network to process the Morgan fingerprint and uses probability as the criterion to decide the most probable reaction rules to apply to the input molecules. Coley et al. \cite{coley2017prediction} also present a synthesis framework that applies all possible reaction templates to reactants, in order to generate all possible candidate reactions and use softmax layer to score and rank all the candidates.

\textbf{Graph-based}.
Unlike Segler and Waller \cite{segler2017neural} who use fingerprints to constitute the reaction template, Coley et al. \cite{coley2017prediction} focus on the reaction cores as the representation. By formalizing reaction prediction as finding missing links between reactant molecules, reaction prediction can also be interpreted as predicting the missing nodes and edges for a given chemical knowledge graph. Based on such understanding, Segler and Waller \cite{segler2017modelling} build a knowledge graph using millions of existing reactions and molecules from the database. Instead of simply constructing molecules as nodes and reactions as edges, they construct both molecules and reactions as nodes and assign different roles to the edges. The corresponding graph matching algorithms can perform product prediction and conditional prediction.

\subsection{Template-free} 

On the other hand, the models without using reaction templates also become prevalent in recent years. When researchers perform the task of reaction prediction, they often first find the reactivity site and then make the comparison to get the final products. Some template-free models mimic this process. 

\textbf{Graph-based}
Jin et al. \cite{jin2017predicting} propose a model based on Weisfeiler-Lehman Network. It takes a two-step strategy similar to the experts with two separate models. The first model is used to find the reaction center by predicting the reactivity scores of pairwise atoms. Based on chemical constraints, the bonds will be generated for those pairs of atoms, possibly the reaction site. Finally, the second model will be applied to rank those candidate products. Later, they improve their work \cite{coley2019graph} by replacing Weisfeiler-Lehman Network with a graph convolution network, then combing reaction center prediction and scoring candidate production as one single task. Qian et al. \cite{qian2020integrating} also propose a two-stage model with a graph convolution network but consider things like hydrogen counts and formal charges when identifying the reaction site. Specifically, they utilize a message passing neural network to get the molecular embedding in each atom and use a convolution-based co-estimation network to predict the reaction site. Finally, they use integer linear programming to search the production space.  

While the above models design in a ``human manner'' with multi-step, there are also end-to-end models. A graph convolution network architecture proposed by Sacha et al. \cite{sacha2021molecule} for both reaction prediction and retrosynthesis generates the reaction by a sequence of graph edit, which is already discussed in the previous section. Focusing on predicting reaction mechanisms instead of the reaction products directly, Bradshaw et al. \cite{bradshaw2018generative} propose an end-to-end generative model called ELECTRO. The reaction with the linear electron flow mechanism is then represented as the sequence of electron steps. Combining reinforcement learning with graphical models, Do et al. \cite{do2019graph} propose a graph transformation policy network, which devotes to tackling the problem using less chemical knowledge. There are three parts in this model: a graph neural network used to represent the input molecules, including reactants and reagents, a node pair prediction network used to output the reaction triples, and a policy network used to generate the intermediate molecules. Bi et al.\cite{bi2021non} proposed Non-autoregressive Electron Redistribution Framework, which views the change of edge as flowing electrons in molecules. Like others, they use graph neural networks to represent the molecule, but instead of predicting the generation or breaking of the molecular bonds, they predict the flow of electrons.

\textbf{String-based}.
Besides the various graphical models, those entirely data-driven sequence-based models are also well-proposed. Schwaller et al. \cite{schwaller2018found} present a model which views reaction prediction as a translation problem and thus applies neural machine translation models. The model consists of two recurrent neural networks. The first is used to encode the sequence representation of the molecule, and the second is a decoder used to produce the outcome products and the corresponding probability. Later, after the advent of the transformer, they further developed a multi-head attention molecular transformer model \cite{schwaller2019molecular}, which takes advantage of the attention mechanism.

Template-based methods have their own unique advantages. The process of finding possible existing reaction rules is similar to the way chemists complete this task. These methods inherently enhance the interpretability of models. The most important parts of template-based models are the generation of templates because the quality and numbers of reaction templates directly influence the model performance. The template generation leads to some problems. Firstly, because those models cannot apply the reaction rules or transformations not covered in the template library in the test stage, the generalization becomes a huge problem. Secondly, some hand-encoding rules and transforms have bias because they are generated from a small part of chemists \cite{wei2016neural} and thus not that accurate. Besides, some early models which use fingerprints as inputs \cite{segler2017neural,coley2017computer,baylon2019enhancing,segler2018planning} have limitation from the data side. Fingerprints do not include much information, especially on the substructures, connectivity \cite{chen2021deep}, and local contexts. In the later stage, the fully data-driven template generating algorithms tackle the second problem, but the computational source limitation problem still exists. It leads to the restriction on the amounts of templates, which further negatively influences the model’s coverage, scalability, and diversity. As Schwaller et al. \cite{schwaller2019molecular} point out in their article, those auto-generating algorithms have flaws logically. Those template extracting algorithms are based on atom mapping, which comes from a template library. This creates a logical ring and is thus not that credible and accurate. 

On the other hand, template-free models overcome the drawbacks mentioned above. Graphical models use graph representation, and it has excellent advantages in interpretability. As the atoms and bonds can be directly converted to nodes and edges, those graphs and subgraphs are explainable chemically. By using graph neural network, the local information can be aggregated to each node and edge. As graph neural network learns task-specific embeddings, for both retrosynthesis and reaction prediction, many graph-based models take two-stage strategies. By modifying the layers and focusing on latent spaces, some models like Chen et al. \cite{ishiguro2020data} also enhance the diversity of the models. Although graphical models have promising outcomes, wonderful interpretability, and scalability, they require atom-mapped datasets in the training stage \cite{wei2016neural}. Similar to the problems for template-based models, they implicitly use pre-defined reaction templates and thus maybe not be as credible as ground truth. Sequence-based models utilize the advantages of transformer and perform well on those tasks; however, it lacks interpretability because the string can hardly expose the local information for the three-dimensional structured molecules. The integration of sequence-based models containing graphical structures as input information \cite{somnath2021learning,gao2022semiretro,yan2020retroxpert,sun2020energy} take advantage of both sides, but it still can not bypass the drawbacks of atom-mapped datasets.
\section{Datasets and Benchmarks}
\label{sec:datasets}
\subsection{Datasets}

\textbf{PubChem}. PubChem \cite{kim2021pubchem} is an open chemistry database at the National Institutes of Health (NIH). Most samples in the database are small molecules, but large molecules such as nucleotides, carbohydrates, lipids, peptides, and chemically-modified macromolecules also exist. The information and annotations of molecules include chemical structures, identifiers, chemical and physical properties, biological activities, patents, health, safety, toxicity data, and many others. PubChem contains 93.9 million entries for compounds (54 million in September 2014), including pure and characterized compounds; 236 million entries for molecules (163 million in September 2014), and also mixtures, extracts, complexes, and uncharacterized substances.

\textbf{ChEMBL}. ChEMBL~\cite{gaulton2012chembl} is a manually curated database of bioactive molecules with drug-like properties. It brings together chemical, bioactivity and genomic data to aid the translation of genomic information into effective new drugs. It contains 14,885 targets and 2,157,379 distinct compounds by February 2022. The annotation of ChEMBL mainly includes the positive and negative effects of the compounds on the corresponding targets.

\textbf{ZINC}. ZINC \cite{sterling2015zinc} is a free database of commercially-available compounds for virtual screening. ZINC contains over 230 million purchasable compounds in ready-to-dock, 3D formats and over 750 million purchasable compounds.

\textbf{GDB}. GDB \cite{fink2005virtual} is the largest publicly available small organic molecule database. GDB dataset is a collection of SMILES strings, GDB-11 \cite{fink2007virtual} has 100 million molecules, GDB-13 \cite{blum2009970} has 1 billion molecules, and GDB-17 \cite{ruddigkeit2012enumeration} has 160 billion molecules. The number in the postscript indicates the maximum number of atoms in the molecule in the dataset.

\textbf{QM9}. QM9 \cite{ramakrishnan2014quantum} is a benchmark dataset that contains 134K stable organic molecules. QM9 records geometric, energetic, electronic, and thermodynamic properties. GEOM-QM9 is a subset of QM9, including 3D geometry structures with small molecular mass and few rotatable bonds. All molecules are modeled based on DFT methods, and same for the calculated quantum mechanical properties. Descriptions of recorded properties are described in TABLE~\ref{tab:property}.

\begin{table}[!t]
\centering
 \renewcommand\arraystretch{1.2}
\caption{Description of properties in QM9 dataset}
\label{tab:property}
\begin{tabular}{l|l}
\toprule
Properties               & Description                             \\ \midrule
$U_0$                    & Internal energy at 0K                   \\
$U$                      & Internal energy at 298.15K              \\ 
$G$                      & Free energy at 298.15K                  \\ 
$H$                      & Enthalpy at 298.15K                     \\ 
$C_v$                    & Heat capacity at 298.15K                \\ 
$\epsilon_{HOMO} (HOMO)$ & Energy of HOMO                          \\ 
$\epsilon_{LUMO} (LUMO)$ & Energy of LUMO                          \\ 
$\epsilon_{gap} (gap)$   & Gap ($\epsilon_{LUMO}-\epsilon_{HOMO}$) \\ 
ZPVE                     & Zero point vibrational energy           \\ 
$<R^2>$                  & Electronic spatial extent               \\ 
$\mu$                    & Dipole moment                           \\ 
$\alpha$                 & Isotropic polarizability                \\ \bottomrule
\end{tabular}
\vspace{-20pt}
\end{table}

\textbf{GEOM-Drugs}. GEOM-Drugs \cite{axelrod2022geom} is a dataset that provides 3D molecular geometries for larger drug molecules, up to a maximum of 181 atoms (91 heavy atoms). It also contains multiple conformations for each molecule, with a larger structure variance.

\textbf{USPTO}. USPTO \cite{schneider2016s} is an open-source chemical reaction database and contains about 3.7 million reactions. The most popular datasets for reaction and retrosynthesis prediction are from USPTO. The frequently used dataset includes USPTO-50K (contains 50k reactions), USPTO-MIT,    USPTO-15k (contains 50k reactions), and USPTO-FULL (contains 1M reactions). Among all of these datasets, the most commonly used one is USPTO-50K. It randomly picks 50000 reactions from US patents, and all reactions are assigned to a specific reaction type. For single-step retrosynthesis, USPTO-50K often serves as a benchmark. However, the distribution of the reaction classes is not that uniform, and some patented syntheses are not validated by experiments and thus may have flaws.

\textbf{Reaxys}. Reaxys \cite{schneider2016s} is a database used by scholars who focus on reaction prediction. It has the world's largest physical and chemical properties, factual reaction database and pharmaceutical chemistry database. This database has 253M substances, 58M reactions, 96M documents, 32M patents, and 42M bioactivities. Despite the rich contents, large scale, and long history of Reaxys, in order to extract useful information from the database and then apply it to the models, researchers need additional efforts in pre-processing (filter) the data.

\textbf{MoleculeNet}. MoleculeNet \cite{wu2018moleculenet} is a data collection specifically designed for evaluating molecular prediction models. The dataset collection includes over 700,000 compounds of various properties, mainly in four categories, quantum mechanics, physical chemistry, biophysics, and physiology. The first two collections are used to test as regression tasks, while the latter two are for classification ones. Quantum mechanics are information on electronic properties determined by DFT-based methods. ESOL, FreeSolv, and Lipophilicity in the physical chemistry category respectively include data of solubility, hydration free energy, and distribution coefficient in water. Physiology collection, including drug-related and toxicology data, with Biophysics collection composed of data of biological properties such as binding affinities and measured biological activities, are the most frequently used datasets for model evaluation in property prediction in the articles above.

\textbf{ExcapeDB}.
ExcapeDB~\cite{sun2017excape} is a comprehensive dataset that integrates active and inactive compounds from PubChem and ChEMBL. It serves as a data hub, providing researchers worldwide with convenient access to a standardized chemogenomics dataset, with the data and accompanying software available under open licenses.

\textbf{LSC}.
LSC~\cite{mayr2018large} is a large benchmark dataset assembled from the ChEMBL database, enabling accurate evaluation of machine learning methods for compound target prediction. With over 500,000 compounds and more than 1,000 assays, the dataset represents a diverse array of target classes, including enzymes, ion channels, and receptors.

\subsection{Evaluation Metrics}
There are multiple evaluation metrics for molecule generation. Novelty illustrates the percentage of generated molecules dissimilar to molecules in the training set~\cite{walters2020assessing}.
Diversity is calculated based on pairwise Tanimoto similarity, which measures the diversity of generated molecules. Property score is the average score of the top 100 molecules. The success rate is the percentage of generated molecules that satisfy all objective functions \cite{olivecrona2017molecular}.
For the 3D molecule generation problem, generated and reference molecule sets are used for evaluation. Coverage (COV) score \cite{xu2021learning} measures the percentage of molecule conformations in one set covered by another. Matching (MAT) score measures the distance of the closest neighbor between two sets. ``-P'' and ``-R'' means precision metric and recall metric, respectively.

Mean-absolute error (MAE), root-mean-square error (RMSE), and ROC-AUC scores are the most commonly used evaluation metrics for molecular property prediction. Performance is measured on either classification or regression tasks, in which QM datasets and datasets from MoleculeNet \cite{wu2018moleculenet} are commonly used for evaluation. If property prediction is regarded as a classification task, which is often the case, models are evaluated on a set of given labels after training. Regression tasks for molecular property prediction are more demanding than the classification ones, requiring a competitive model to predict the exact numeric values of molecular properties \cite{wang2022molecular}. Supervised models are commonly introduced in the evaluation process as baselines.

\begin{table*}[h!]
\centering
\caption{Datasets and Evaluation Metrics for Selected Models}
\label{table two}
\setlength{\tabcolsep}{1mm}{
\begin{tabular}{ccccccccccc}
   \toprule
   Task and Dataset & Model&Year & \multicolumn{8}{c}{Evaluation Metrics} \\
   \toprule
  \textbf{2D Molecule Generation} & & &\multicolumn{2}{c}{Diversity ↑}&\multicolumn{2}{c}{Property Score ↑}&\multicolumn{2}{c}{Success Rate ↑}\\
  
   \midrule
    ZINC
   &GCPN \cite{you2018graph}&2018 &\multicolumn{2}{c}{0.596}& \multicolumn{2}{c}{0.450}&\multicolumn{2}{c}{-}\\
   &MolDQN \cite{zhou2019optimization}&2019 &\multicolumn{2}{c}{0.597}& \multicolumn{2}{c}{0.365}&\multicolumn{2}{c}{-}\\
    &LSTM \cite{segler2018generating}&2019& \multicolumn{2}{c}{0.706}& \multicolumn{2}{c}{0.672}&\multicolumn{2}{c}{-}\\
   &BOSS \cite{moss2020boss}&2020 &\multicolumn{2}{c}{0.561} &\multicolumn{2}{c}{0.504}&\multicolumn{2}{c}{-}\\
   &ChemBO \cite{korovina2020chembo}&2020 &\multicolumn{2}{c}{0.701}& \multicolumn{2}{c}{0.648}&\multicolumn{2}{c}{-}\\
   &DST \cite{fu2021differentiable}&2022&\multicolumn{2}{c}{\textbf{0.755}}&\multicolumn{2}{c}{\textbf{0.752}}&\multicolumn{2}{c}{-}\\
   
   \midrule
    ChEMBL
   &REINVENT \cite{olivecrona2017molecular}&2017&\multicolumn{2}{c}{0.666}&\multicolumn{2}{c}{-}&\multicolumn{2}{c}{46.6\%}\\
   &JT-VAE \cite{jin2018junction}&2018& \multicolumn{2}{c}{0.277}&\multicolumn{2}{c}{-}&\multicolumn{2}{c}{5.4\%}\\
   &GA+D \cite{nigam2019augmenting}&2019& \multicolumn{2}{c}{0.363}&\multicolumn{2}{c}{-}&\multicolumn{2}{c}{85.7\%}\\
    &RationaleRL \cite{jin2020multi}&2020& \multicolumn{2}{c}{0.706}&\multicolumn{2}{c}{-} &\multicolumn{2}{c}{75.0\%}\\
   &MARS \cite{xie2021mars}&2021&\multicolumn{2}{c}{\textbf{0.719}}&\multicolumn{2}{c}{-}&\multicolumn{2}{c}{92.3\%}\\
   &MolEvol \cite{chen2021molecule}&2021&\multicolumn{2}{c}{0.681}&\multicolumn{2}{c}{-}&\multicolumn{2}{c}{\textbf{93.0\%}}\\

   \midrule
   \multirow{2}{*}{\textbf{3D Molecule Generation}}&&&\multicolumn{2}{c}{COV-R (\%) ↑}&\multicolumn{2}{c}{MAT-R (Å) ↓}&\multicolumn{2}{c}{COV-P (\%) ↑}&\multicolumn{2}{c}{MAT-P (\%) ↓}\\
   &&&Mean&Median&Mean&Median&Mean&Median&Mean&Median\\
   
   \midrule
   GEOM-Drugs
   &CVGAE \cite{mansimov2019molecular}&2019&0.00& 0.00 &3.0702& 2.9937 &- &- &-& -\\
   &GRAPHDG \cite{simm2019generative}&2020&8.27 &0.00 &1.9722 &1.9845& 2.08 &0.00& 2.4340 &2.4100\\
   &CGCF \cite{xu2021learning}&2021&53.96& 57.06& 1.2487& 1.2247 &21.68 &13.72 &1.8571 &1.8066\\
   &CONFVAE \cite{xu2021end}&2021&55.20& 59.43 &1.2380 &1.1417 &22.96& 14.05 &1.8287 &1.8159\\
   &CONFGF \cite{shi2021learning}&2021&62.15 &70.93 &1.1629 &1.1596& 23.42 &15.52 &1.7219 &1.6863\\
   &GEOMOL \cite{zhu2022direct}&2022&67.16 &71.71& 1.0875 &1.0586 &-& -& - &-\\
   &GEODIFF \cite{xu2022geodiff}&2022&\textbf{89.13}& \textbf{97.88} &\textbf{0.8629} &\textbf{0.8529} &\textbf{61.47} &\textbf{64.55} &\textbf{1.1712} &\textbf{1.1232}\\
   \midrule
   GEOM-QM9
   &CVGAE \cite{mansimov2019molecular}&2019&0.09& 0.00& 1.6713 &1.6088& -& -& -& -\\
   &GRAPHDG \cite{simm2019generative}&2020&3.33 &84.21 &0.4245 &0.3973 &43.90 &35.33 &0.5809 &0.5823\\
   &CGCF \cite{xu2021learning}&2021&78.05& 82.48& 0.4219 &0.3900 &36.49& 33.57 &0.6615& 0.6427\\
   &CONFVAE \cite{xu2021end}&2021&77.84& 88.20& 0.4154 &0.3739 &38.02& 34.67& 0.6215& 0.6091\\
   &GEOMOL \cite{zhu2022direct}&2022&71.26& 72.00& 0.3731 &0.3731& -& -& -& -\\
   &CONFGF \cite{shi2021learning}&2021&88.49 &\textbf{94.31}& 0.2673 &0.2685& 46.43 &43.41 &0.5224 &0.5124\\
   &GEODIFF \cite{xu2022geodiff}&2022&\textbf{90.07} &93.39 &\textbf{0.2090 }&\textbf{0.1988} &\textbf{52.79} &\textbf{50.29}& \textbf{0.4448}& \textbf{0.4267}\\
   
    \midrule
   \multirow{2}{*}{\textbf{Retrosynthesis Prediction}}&&&\multicolumn{8}{c}{Accuracies (\%) ↑}\\
   &&&\multicolumn{2}{c}{Top-1}&\multicolumn{2}{c}{Top-3}&\multicolumn{2}{c}{Top-5}&\multicolumn{2}{c}{Top-10}\\
   
   \midrule
   USPTO-50k
   &SCROP \cite{zheng2019predicting}&2019&\multicolumn{2}{c}{59.0}& \multicolumn{2}{c}{74.8} & \multicolumn{2}{c}{78.1} & \multicolumn{2}{c}{81.1}  \\
   &GLN \cite{dai2019retrosynthesis}&2019& \multicolumn{2}{c}{63.2} & \multicolumn{2}{c}{77.5} & \multicolumn{2}{c}{83.4} & \multicolumn{2}{c}{89.1}\\
   &G2Gs \cite{shi2020graph}&2020&\multicolumn{2}{c}{61.0} & \multicolumn{2}{c}{81.3} & \multicolumn{2}{c}{86.0} & \multicolumn{2}{c}{88.7}\\
   &RetroXpert \cite{yan2020retroxpert}&2020& \multicolumn{2}{c}{\textbf{70.4}} & \multicolumn{2}{c}{83.4} & \multicolumn{2}{c}{85.3} & \multicolumn{2}{c}{86.8} \\
   &Dual \cite{sun2020energy}&2020&\multicolumn{2}{c}{65.7}&\multicolumn{2}{c}{81.9} &\multicolumn{2}{c}{84.7}& \multicolumn{2}{c}{85.9}\\
   &MEGAN \cite{sacha2021molecule}&2021& \multicolumn{2}{c}{60.7} &\multicolumn{2}{c}{82.0} & \multicolumn{2}{c}{87.5} & \multicolumn{2}{c}{91.6}\\
   &GraphRetro \cite{somnath2021learning}&2021&\multicolumn{2}{c}{63.9}&\multicolumn{2}{c}{81.5}&\multicolumn{2}{c}{85.2}&\multicolumn{2}{c}{88.1}\\
   &LocalRetro \cite{chen2021deep}&2021&\multicolumn{2}{c}{63.9}&\multicolumn{2}{c}{\textbf{86.8}}&\multicolumn{2}{c}{\textbf{92.4}}&\multicolumn{2}{c}{\textbf{96.3}}\\
   &Retroformer \cite{wan2022retroformer}&2022&\multicolumn{2}{c}{64.0}&\multicolumn{2}{c}{82.5} &\multicolumn{2}{c}{86.7}&\multicolumn{2}{c}{90.2}\\
   
   \midrule
   USPTO-full
   &RetroSim \cite{coley2017computer}&2017& \multicolumn{2}{c}{32.8} &\multicolumn{2}{c}{-} & \multicolumn{2}{c}{-} & \multicolumn{2}{c}{56.1}\\
   &MEGAN \cite{sacha2021molecule}&2021& \multicolumn{2}{c}{33.6} &\multicolumn{2}{c}{-} & \multicolumn{2}{c}{-} & \multicolumn{2}{c}{63.9}\\
   &GLN \cite{dai2019retrosynthesis}&2019& \multicolumn{2}{c}{39.3} & \multicolumn{2}{c}{-} & \multicolumn{2}{c}{-} & \multicolumn{2}{c}{63.7}\\
   &GTA \cite{seo2021gta}&2021& \multicolumn{2}{c}{46.6} & \multicolumn{2}{c}{-} & \multicolumn{2}{c}{-} & \multicolumn{2}{c}{70.4}\\
   &Reaction-Aware \cite{zhao2022leveraging}&2022& \multicolumn{2}{c}{\textbf{51.6}} & \multicolumn{2}{c}{-} & \multicolumn{2}{c}{-} & \multicolumn{2}{c}{\textbf{70.7}}\\
    \midrule
   \multirow{2}{*}{\textbf{Reaction Prediction}}
   &&&\multicolumn{3}{c}{Coverages (\%) ↑} && \multicolumn{4}{c}{Accuracies (\%) ↑} \\
   &&&  C@6 &   C@8 &   C@10   && Top-1 & Top-2 & Top-3 & Top-5 \\
   
   \midrule
   USPTO-15k
   &WLDN \cite{jin2017predicting}&2017&81.6 & 86.1 & 89.1 && - & - & -& -\\
   &GTPN \cite{do2019graph}&2019&\textbf{88.9} & \textbf{92.0} & \textbf{93.6} && - & - & -& - \\
   \midrule
   USPTO-MIT
   &WLDN \cite{jin2017predicting}&2017&-& - & - && 79.6 & - & 87.7& 89.2\\
   &GTPN \cite{do2019graph}&2019&-&-&- && 83.2 & - & 86.0& 86.5\\
   &Molecular Transformer\cite{schwaller2019molecular}&2019&-&-&-&&90.4&\textbf{93.7}&\textbf{94.6}&95.3\\
   &MEGAN \cite{sacha2021molecule}&2021&-&-&-&&89.3&92.7&94.4&\textbf{95.6}\\
   &NERF \cite{bi2021non}&2021&-&-&-&&\textbf{90.7} & 92.3 &93.3 & 93.7\\
   
   \bottomrule
\end{tabular}}
\vspace{-10pt}
\end{table*}

\begin{table*}[h!]
\centering
\caption{Datasets and Evaluation Metrics for Selected Models}
\label{table three}
\begin{tabular}{ccccccccc}
   \toprule
   Task and Dataset & Model&Year & \multicolumn{4}{c}{Evaluation Metrics} \\
   
   \midrule
   \multirow{2}{*}{\textbf{Property Prediction}}&&&\multicolumn{6}{c}{MAE ↓}\\
   &&&$U_0$(eV)&$U$(eV)&$G$(eV)&$H$(eV)&$C_v$($\frac{cal}{mol K}$)&$HOMO$(eV)\\
   \midrule
   QM9
   &InfoGraph \cite{sun2019infograph}&2019 &0.1410&0.1702 & 0.1592 & 0.1552 & 0.1965 & 0.1605 \\
   &MGCN \cite{lu2019molecular}&2019 &0.0129&0.0144 & 0.0146 & 0.0162 & 0.0380 & 0.0421 \\
   &ASGN \cite{hao2020asgn}&2020 &0.0562&0.0594 & 0.0560 & 0.0583 & 0.0984 & 0.1190 \\
   &SphereNet \cite{liu2021spherical}&2021 &\textbf{0.0062}&\textbf{0.0064} & \textbf{0.0078} & \textbf{0.0063} & \textbf{0.0215} & \textbf{0.0228} \\
   
   \midrule
    \multirow{2}{*}{\textbf{}}&&&\multicolumn{6}{c}{MAE ↓}\\
   &&&$LUMO$(eV)&$gap$(eV)&$ZPVE$(eV)&$R^2$ ($a_0^2$)&$\mu$(D)&$\alpha$ ($a_0^3$)\\
   \midrule
    QM9
   &InfoGraph \cite{sun2019infograph}&2019 &0.1659&0.2421 & 0.00036 & 4.92 & 0.3168 & 0.5444 \\
   &MGCN \cite{lu2019molecular}&2019 &0.0574&0.0642 & 0.00112 & \textbf{0.11} & 0.0560& \textbf{0.0300} \\
   &ASGN \cite{hao2020asgn}&2020 &0.1061&0.2012 & \textbf{0.00017} & 1.38 &  0.1947 & 0.2818 \\
   &SphereNet \cite{liu2021spherical}&2021 &\textbf{0.0189}&\textbf{0.0311}& 0.00110 & 0.27 & \textbf{0.0245} & 0.0449 \\
   
   \midrule
    \multirow{2}{*}{\textbf{}}&&&\multicolumn{6}{c}{MAE ↓}\\
   &&&FreeSolv&ESOL&Lipo&QM7&QM8\\
   \midrule
   MoleculeNet
   &MPNN \cite{gilmer2017neural}&2017 &2.185 &1.167 & 0.672 & 113.0 & 0.0150 \\
   &MGCN \cite{lu2019molecular}&2019 &3.349& 1.266 & 1.113 & 77.6 & 0.0220 \\
   &GROVER \cite{rong2020self}&2020 &1.544&0.831 & 0.560 & 72.6 & 0.0125 \\
   &multiple SMILES\cite{li2022novel}&2022 &\textbf{0.786} &\textbf{0.445} & \textbf{0.548} & - & - \\
   &ChemNet \cite{yang2021deep}&2021 &- &- & - & 60.1 & \textbf{0.0100} \\
   &PhysChem \cite{yang2021deep}&2021 &- &- & - & \textbf{59.6} & 0.0101 \\

   \midrule
    \multirow{2}{*}{\textbf{}}&&&\multicolumn{6}{c}{RMSE ↓}\\
   &&&FreeSolv&ESOL&Lipo\\
   \midrule
   MoleculeNet
   &N-Gram \cite{liu2019n}&2019&2.51& 1.10& 0.88\\
   &GROVER \cite{rong2020self}&2020& 2.48& \textbf{0.99}& 0.66\\ 
   &MolCLR \cite{wang2022molecular}&2022 &\textbf{2.20} &1.11 & \textbf{0.65} \\
   
   \midrule
    \multirow{2}{*}{\textbf{}}&&&\multicolumn{6}{c}{ROC-AUC ↑}\\
    &&&BBBP&SIDER&ClinTox&BACE&Tox21&ToxCast\\
   \midrule
   MoleculeNet
   &AttrMasking \cite{hu2019strategies}&2019 &0.702 &0.604 & 0.686 & 0.772 & 0.742 & 0.625 \\
   &ContextPred \cite{hu2019strategies}&2019 &0.712 & 0.593 &0.737 & 0.786 & 0.733 & 0.628 \\
   &InfoGraph \cite{sun2019infograph}&2019 &0.692&0.592 & 0.751 & 0.739 & 0.730 & 0.620 \\
   &GraphCL \cite{you2020graph}&2020 &0.675 &0.601 & 0.789 & 0.687 & 0.750 & 0.628 \\
   &GROVER \cite{rong2020self}&2020 &0.718&0.637 & \textbf{0.843} &\textbf{ 0.822} & \textbf{0.765} & \textbf{0.635} \\
   &ChemBERTa \cite{ adilov2021generative}&2021 &0.643&- & 0.733 & - & 0.728 & - \\
   &GraphMVP \cite{liu2021pre}&2021 &\textbf{0.724}&\textbf{0.639} &0.791& 0.812 & 0.759 & 0.631 \\

    \midrule
    \multirow{2}{*}{\textbf{}}&&&\multicolumn{6}{c}{AUC ↑}\\
     &&&ClinTox&Tox21\\
   \midrule
   MoleculeNet
   &Chemception \cite{goh2017chemception}&2017 &0.745 &0.766\\
   &SMILES2vec \cite{goh2017smiles2vec}&2017 &0.693 &0.810\\
   &CheMixNet \cite{paul2018chemixnet}&2018 &0.944&\textbf{0.920}\\
   &TOP \cite{peng2019top}&2019 &\textbf{0.946} &-\\

   \bottomrule
\end{tabular}
\vspace{-10pt}
\end{table*}

The standard evaluation metric for retrosynthesis and reaction prediction is the top-k accuracy, a concept from probability. This metric will compute the percentage of times the correct answer is in the highest k ranking scores. Referring to the application in retrosynthesis prediction and reaction prediction, the match of the ground-truth reactants (or products) with the predicted one will be computed. Based on such an exact match, they will then apply top-k accuracy. The choice of k varies from 1 to 50, depending on different works. Another evaluation metric used in reaction prediction is Coverage@k, which is the proportion of reactions in all the actual atom pairs founded in the set of the predicted atom pairs.

\subsection{Benchmark Analysis}
According to evaluation data of 2D molecule generation models in table~\ref{table two}, we could conclude that graph models like DST \cite{fu2021differentiable}, MARS \cite{xie2021mars} and MolEvol \cite{chen2021molecule} have better performance than SMILES models like GA+D \cite{nigam2019augmenting} and LSTM \cite{brown2019guacamol}. This may be because SMILES strings fail to represent distances between atoms. For example, atoms close in the SMILES string may be far apart in the actual structure. Meanwhile, models using the notion of motif like MolEvol \cite{chen2021molecule}, MARS \cite{xie2021mars} and RationaleRL \cite{jin2020multi} have shown better success rate than atom by atom generation models. However, novelty of these models is limited since the motifs are extracted from the training set. For 3D molecule generation, variational auto-encoder and flow are two commonly used techniques in previous works. GEODIFF \cite{xu2022geodiff} combines the diffusion model with the molecular generation task and achieves the state-of-the-art result. Thus, finding models that better match 3D molecular invariant features is challenging for the 3D molecule generation task. 

Comparing the model performances on the QM9 datasets, we find out that utilizing unlabeled data appears to be an effective solution for predicting a large number of molecules \cite{hao2020asgn,lu2019molecular}. ASGN improves the transferability by weight transfer, while MGCN transfers the knowledge of small molecules to large ones to resolve the structural shortage of the dataset. These robust models achieve a more than 50$\%$ reduction on properties, such as, $U_0$, $C_v$, $\alpha$ compared with baselines that rely on limited labeled data, showing a great necessity to improve generalizability and transferability. Incorporating multilevel information, especially the spatial position \cite{lu2019molecular,liu2021spherical}, are proved to contribute to better performance. Instead of simply putting 3D information, such as angle, into use, spherical message passing (SMP) \cite{liu2021spherical} produces accurate, complete, and physically meaningful data representations on 3D graphs. The advancement has shown a significant improvement in almost all properties.

As discovered in the molecule generation task, the overall performances on the MoleculeNet dataset reveal the overwhelming advantages of graph models over the string ones, especially for those involving multilevel structural details \cite{rong2020self}. The superb performance of TOP \cite {peng2019top} on the ClinTox dataset demonstrates the strengths of incorporating multiple kinds of fingerprints compared to SMILES2vec \cite{goh2017smiles2vec}. The specially-designed mixed representation with SMILES strings and physiochemical properties demonstrate its superior capability for toxicity prediction on the ClinTox dataset. The innovation of combining properties with learned structural features provides insights to overcome shortcomings of existing explorations.

The overall model performance on the MoleculeNet dataset also indicates that self-supervised pre-training tasks result in a performance boost. While supervised pre-training tasks introduce the risk of negative transfer \cite{hu2019strategies}, well-designed self-supervised pre-trained models show rival performance in learning implicit domain knowledge \cite{rong2020self,you2020graph,wang2022molecular}. However, some tasks that involve alteration of the molecule structure, such as subgraph removal \cite{wang2022molecular}, result in poor performance in datasets sensitive to topology changes, such as the BBBP dataset. Moreover, the composition of different augmentation strategies is not bound to perform better since removing various substructures might easily lead to the loss of important structural information. Furthermore, models involving 3D information have superior performance on datasets of quantum mechanical properties. MGCN \cite{lu2019molecular}, though not showing distinct advantages on other datasets, shows a competitive result as those pre-training techniques on QM7 and QM8 datasets.

For retrosynthesis prediction, RetroXpert \cite{yan2020retroxpert} and Reaction-Aware \cite{zhao2022leveraging} combine the information from both graph representation and sequence representation and perform well. Generally speaking, template-free models have better performance. However, LocalRetro \cite{chen2021deep}, a template-based model, outperforms others. This may be due to the model design, which not only strictly constrains the templates but also considers global structure. For reaction prediction, models such as GTPN \cite{do2019graph}, Molecular Transformer \cite{schwaller2019molecular} and NERF \cite{bi2021non} which utilize the state-of-art models in other applications of machine learning also get good results in this task. 
\section{Challenges and Future trend}
\label{sec:discussion}

\subsection{Challenges}

\textbf{Data}.
Despite all the machine learning models discussed above, it should be emphasized that data is the limiting factor in drug discovery. Data must be of sufficient quantity and high quality to make the model useful and practical. Although existing databases contain a large number of molecules, the number of data for a certain task can be very sparse \cite{stanley2021fs,altae2017low,subramanian2016computational}. For instance, the training data is particularly limited for generating polymers with given properties \cite{guo2022data}. Meanwhile, the quality of datasets still needs to be improved. The USPTO datasets are the most popular datasets for retrosynthesis due to their integrity and accessibility. However,
the USPTO dataset still suffers from atom-mapping inaccuracy, and noisy stereochemical data \cite{chen2020retro}. Data inconsistency is also a crucial problem in drug design~\cite{sidorov2019predicting}. For example, data inconsistencies from different centres lead to substantial inconsistencies in drug responses in large pharmacogenomic studies~\cite{safikhani2016revisiting}.

\textbf{Evaluation Metrics}.
The currently used evaluation metrics need to be improved. 
Renz et al. \cite{renz2019failure} have shown that it is easy to maximize ``validity'' by inserting a carbon atom into the training SMILES string. Current evaluation metrics for retrosynthesis also have misleading meanings \cite{schwaller2020predicting,thakkar2020datasets,fortunato2020data}. Therefore, although a large number of evaluation metrics have been provided, these metrics are simplified due to computational complexity and differ from the real definition to some extent. How to balance computational complexity with real definitions is still an unsolved problem, which requires continuous efforts to refine the evaluation methods better.

\textbf{Interpretability}. 
Interpretability issue is another challenge in drug discovery. Although existing models perform well, the reasons behind them are still unclear. For instance, retrosynthesis can predict the initial products. However, if the mechanism behind retrosynthesis could also be provided, it would be helpful for the research of chemical reactions. More specifically, there are four aspects to cover \cite{jimenez2020drug}: 1) Transparency, which is knowing how the system reaches a particular answer; 2) Justification, which is elucidating why the answer provided by the model is acceptable; 3) Informativeness, which is providing new information to human decision-makers; and 4) Uncertainty estimation, which is quantifying how reliable a prediction is. An interpretable model has many advantages. It can help the expert to analyze the causes when the performance of model is poor and provide directions for further improvement. When the interpretation of the model is consistent with the prior knowledge, the confidence of the model can be improved \cite{vamathevan2019applications}. Although there are some explorations to the interpretability via case studies \cite{fu2021differentiable,preuer2019interpretable,du2022interpretable}, how to quantitatively evaluate interpretability remains a challenge \cite{du2022molgensurvey}.

\textbf{Uncertainty Quantification}. 
Most work has focused on improving model accuracy, while quantifying uncertainty requires more attention~\cite{mervin2021uncertainty}. Model accuracy can only represent the reliability of the entire model. However, uncertainty quantification predicts the property interval of a given molecule at a certain confidence level, estimating reliability at the instance level~\cite{hernandez2022conformal}.
By combining the prediction results of the model with the quantification of uncertainty, it is possible to screen for reliable molecules with good properties, thus improving the efficiency of drug design. In addition, the algorithm using uncertainty quantization is robust when exploring the domain outside the distribution of training data, which is important for the model to explore in the huge and diverse chemical space~\cite{hie2020leveraging}.

\subsection{Future Trend}

For \textbf{molecule generation} tasks, the molecular representation framework has gradually shifted from SMILES sequences to graph neural networks. Compared with SMILES sequences, graph neural networks can better represent molecular structures, especially cyclic ones. However, graph neural networks can only assign atom relationships to nodes and edges on the graph. Combining the constraints between atoms with neural networks remains to be explored. Meanwhile, this survey focuses on structured small molecule generation. Other compounds such as protein, gene and crystal with more complex data structures may also benefit from the methods discussed here.

For \textbf{molecular property prediction} tasks, the current trend of leveraging 2D topology and 3D geometry will be sustained. It appears that mutual information maximization and the MPNN framework will be continued to act as the basic theme. While most architectures focus on molecular representations, transferability to downstream tasks requires more attention to avoid the ``negative transfer'' effect since molecular property prediction is often viewed as a downstream task. Novel pre-training strategies will continue to be a powerful trend since these strategies address the limited labeled data and out-of-distribution prediction problems to some extent. Novel strategies for combining molecular properties, such as chemical reactions and molecular dynamics, will be explored with a specific focus on multi-dimensional structures.

For \textbf{retrosynthesis and reaction prediction} tasks, the methods for both tasks can be classified into two categories, \textit{i.e.}, template-based and template-free. The development space is limited for the multi-step template-based models due to the weak generalization ability. Meanwhile, more and more template-free models are proposed because of their excellent coverage, scalability, and diversity. 

However, most of the proposed methods lack information about conditions, such as reagents, catalysts, solvents, temperature. Although recent developed algorithms have made a preliminary attempt on this aspect \cite{schwaller2020predicting,vaucher2021inferring,wang2020towards}, the conversion of algorithms to experimental procedures should receive more attention in the future \cite{vaucher2021inferring}.

Although models and algorithms are proposed separately for different tasks, those tasks have intrinsic relationships. For example, after a molecule is produced by generation models, experts need to know some basic properties of this molecule (\textit{e.g.}, whether it is toxic, whether it can be absorbed by human body) as well as how to synthesize this molecule. This is where the property prediction and retrosynthesis algorithms come into play. Thus, making connections between different tasks and integrating them might be a trend in the future. Furthermore, most of the existing models are based on theory and lack experimental verification. The results of the model have a guiding effect on the experiment, and the data obtained from the experiment can also be fed back to improve the model. The potential for closing the loop for theoretical model and practical experiments is quite profound \cite{henson2018designing}.

\section{Conclusion}
\label{sec:conclusion}
In this survey, we comprehensively review molecule generation, molecular property prediction, retrosynthesis, and reaction prediction tasks under the setting of deep learning. 
 We categorize the existing models based on the molecule representation they employ. Additionally, we delve into the datasets and evaluation metrics and analyze benchmark models in terms of performance. Finally, we highlight the challenges and outline the future trend of drug discovery utilizing deep learning methods.




\section*{Acknowledgments}
This work is supported by National Natural Science Foundation of China (62106219) and Natural Science Foundation of Zhejiang Province (QY19E050003).

\bibliographystyle{IEEEtran}
\bibliography{ref} 

\begin{thebibliography}{100}
\providecommand{\url}[1]{#1}
\csname url@samestyle\endcsname
\providecommand{\newblock}{\relax}
\providecommand{\bibinfo}[2]{#2}
\providecommand{\BIBentrySTDinterwordspacing}{\spaceskip=0pt\relax}
\providecommand{\BIBentryALTinterwordstretchfactor}{4}
\providecommand{\BIBentryALTinterwordspacing}{\spaceskip=\fontdimen2\font plus
\BIBentryALTinterwordstretchfactor\fontdimen3\font minus
  \fontdimen4\font\relax}
\providecommand{\BIBforeignlanguage}[2]{{%
\expandafter\ifx\csname l@#1\endcsname\relax
\typeout{** WARNING: IEEEtran.bst: No hyphenation pattern has been}%
\typeout{** loaded for the language `#1'. Using the pattern for}%
\typeout{** the default language instead.}%
\else
\language=\csname l@#1\endcsname
\fi
#2}}
\providecommand{\BIBdecl}{\relax}
\BIBdecl

\bibitem{deng2022artificial}
J.~Deng, Z.~Yang, I.~Ojima, D.~Samaras, and F.~Wang, ``Artificial intelligence
  in drug discovery: applications and techniques,'' \emph{Briefings in
  Bioinformatics}, vol.~23, no.~1, p. bbab430, 2022.

\bibitem{schwaller2019molecular}
P.~Schwaller, T.~Laino, T.~Gaudin, P.~Bolgar, C.~A. Hunter, C.~Bekas, and A.~A.
  Lee, ``Molecular transformer: a model for uncertainty-calibrated chemical
  reaction prediction,'' \emph{ACS central science}, vol.~5, no.~9, pp.
  1572--1583, 2019.

\bibitem{kumar2012fragment}
A.~Kumar, A.~Voet, and K.~Zhang, ``Fragment based drug design: from
  experimental to computational approaches,'' \emph{Current medicinal
  chemistry}, vol.~19, no.~30, pp. 5128--5147, 2012.

\bibitem{sheng2013fragment}
C.~Sheng and W.~Zhang, ``Fragment informatics and computational fragment-based
  drug design: an overview and update,'' \emph{Medicinal Research Reviews},
  vol.~33, no.~3, pp. 554--598, 2013.

\bibitem{sheridan1995using}
R.~P. Sheridan and S.~K. Kearsley, ``Using a genetic algorithm to suggest
  combinatorial libraries,'' \emph{Journal of Chemical Information and Computer
  Sciences}, vol.~35, no.~2, pp. 310--320, 1995.

\bibitem{schwalbe2020generative}
D.~Schwalbe-Koda and R.~G{\'o}mez-Bombarelli, ``Generative models for automatic
  chemical design,'' in \emph{Machine Learning Meets Quantum Physics}.\hskip
  1em plus 0.5em minus 0.4em\relax Springer, 2020, pp. 445--467.

\bibitem{svetnik2003random}
V.~Svetnik, A.~Liaw, C.~Tong, J.~C. Culberson, R.~P. Sheridan, and B.~P.
  Feuston, ``Random forest: a classification and regression tool for compound
  classification and qsar modeling,'' \emph{Journal of chemical information and
  computer sciences}, vol.~43, no.~6, pp. 1947--1958, 2003.

\bibitem{zhang2008qsar}
L.~Zhang, H.~Zhu, T.~I. Oprea, A.~Golbraikh, and A.~Tropsha, ``Qsar modeling of
  the blood--brain barrier permeability for diverse organic compounds,''
  \emph{Pharmaceutical research}, vol.~25, no.~8, pp. 1902--1914, 2008.

\bibitem{frohlich2006kernel}
H.~Fr{\"o}hlich, J.~K. Wegner, F.~Sieker, and A.~Zell, ``Kernel functions for
  attributed molecular graphs--a new similarity-based approach to adme
  prediction in classification and regression,'' \emph{QSAR \& Combinatorial
  Science}, vol.~25, no.~4, pp. 317--326, 2006.

\bibitem{chen2019deep}
D.~Chen, S.~Liu, P.~Kingsbury, S.~Sohn, C.~B. Storlie, E.~B. Habermann, J.~M.
  Naessens, D.~W. Larson, and H.~Liu, ``Deep learning and alternative learning
  strategies for retrospective real-world clinical data,'' \emph{NPJ digital
  medicine}, vol.~2, no.~1, pp. 1--5, 2019.

\bibitem{borisov2021deep}
V.~Borisov, T.~Leemann, K.~Se{\ss}ler, J.~Haug, M.~Pawelczyk, and G.~Kasneci,
  ``Deep neural networks and tabular data: A survey,'' \emph{arXiv preprint
  arXiv:2110.01889}, 2021.

\bibitem{li2021machine}
H.~Li, K.-H. Sze, G.~Lu, and P.~J. Ballester, ``Machine-learning scoring
  functions for structure-based virtual screening,'' \emph{Wiley
  Interdisciplinary Reviews: Computational Molecular Science}, vol.~11, no.~1,
  p. e1478, 2021.

\bibitem{bomane2019paclitaxel}
A.~Bomane, A.~Gon{\c{c}}alves, and P.~J. Ballester, ``Paclitaxel response can
  be predicted with interpretable multi-variate classifiers exploiting
  dna-methylation and mirna data,'' \emph{Frontiers in genetics}, vol.~10, p.
  1041, 2019.

\bibitem{grinsztajn2022tree}
L.~Grinsztajn, E.~Oyallon, and G.~Varoquaux, ``Why do tree-based models still
  outperform deep learning on tabular data?'' \emph{arXiv preprint
  arXiv:2207.08815}, 2022.

\bibitem{paul2019property}
A.~Paul, A.~Furmanchuk, W.-k. Liao, A.~Choudhary, and A.~Agrawal, ``Property
  prediction of organic donor molecules for photovoltaic applications using
  extremely randomized trees,'' \emph{Molecular informatics}, vol.~38, no.
  11-12, p. 1900038, 2019.

\bibitem{rester2008virtuality}
U.~Rester, ``From virtuality to reality-virtual screening in lead discovery and
  lead optimization: a medicinal chemistry perspective.'' \emph{Current opinion
  in drug discovery \& development}, vol.~11, no.~4, pp. 559--568, 2008.

\bibitem{rollinger2008virtual}
J.~M. Rollinger, H.~Stuppner, and T.~Langer, ``Virtual screening for the
  discovery of bioactive natural products,'' \emph{Natural compounds as drugs
  Volume I}, pp. 211--249, 2008.

\bibitem{ghislat2021recent}
G.~Ghislat, T.~Rahman, and P.~J. Ballester, ``Recent progress on the
  prospective application of machine learning to structure-based virtual
  screening,'' \emph{Current opinion in chemical biology}, vol.~65, pp. 28--34,
  2021.

\bibitem{wallach2015atomnet}
I.~Wallach, M.~Dzamba, and A.~Heifets, ``Atomnet: a deep convolutional neural
  network for bioactivity prediction in structure-based drug discovery,''
  \emph{arXiv preprint arXiv:1510.02855}, 2015.

\bibitem{feng2018padme}
Q.~Feng, E.~Dueva, A.~Cherkasov, and M.~Ester, ``Padme: A deep learning-based
  framework for drug-target interaction prediction,'' \emph{arXiv preprint
  arXiv:1807.09741}, 2018.

\bibitem{davis2011comprehensive}
M.~I. Davis, J.~P. Hunt, S.~Herrgard, P.~Ciceri, L.~M. Wodicka, G.~Pallares,
  M.~Hocker, D.~K. Treiber, and P.~P. Zarrinkar, ``Comprehensive analysis of
  kinase inhibitor selectivity,'' \emph{Nature biotechnology}, vol.~29, no.~11,
  pp. 1046--1051, 2011.

\bibitem{metz2011navigating}
J.~T. Metz, E.~F. Johnson, N.~B. Soni, P.~J. Merta, L.~Kifle, and P.~J. Hajduk,
  ``Navigating the kinome,'' \emph{Nature chemical biology}, vol.~7, no.~4, pp.
  200--202, 2011.

\bibitem{tang2014making}
J.~Tang, A.~Szwajda, S.~Shakyawar, T.~Xu, P.~Hintsanen, K.~Wennerberg, and
  T.~Aittokallio, ``Making sense of large-scale kinase inhibitor bioactivity
  data sets: a comparative and integrative analysis,'' \emph{Journal of
  Chemical Information and Modeling}, vol.~54, no.~3, pp. 735--743, 2014.

\bibitem{gao2018interpretable}
K.~Y. Gao, A.~Fokoue, H.~Luo, A.~Iyengar, S.~Dey, P.~Zhang \emph{et~al.},
  ``Interpretable drug target prediction using deep neural representation.'' in
  \emph{IJCAI}, vol. 2018, 2018, pp. 3371--3377.

\bibitem{yang2019molecular}
Q.~Yang, V.~Sresht, P.~Bolgar, X.~Hou, J.~L. Klug-McLeod, C.~R. Butler
  \emph{et~al.}, ``Molecular transformer unifies reaction prediction and
  retrosynthesis across pharma chemical space,'' \emph{Chemical
  communications}, vol.~55, no.~81, pp. 12\,152--12\,155, 2019.

\bibitem{dong2022deep}
J.~Dong, M.~Zhao, Y.~Liu, Y.~Su, and X.~Zeng, ``Deep learning in retrosynthesis
  planning: datasets, models and tools,'' \emph{Briefings in Bioinformatics},
  vol.~23, no.~1, p. bbab391, 2022.

\bibitem{cook2012computer}
A.~Cook, A.~P. Johnson, J.~Law, M.~Mirzazadeh, O.~Ravitz, and A.~Simon,
  ``Computer-aided synthesis design: 40 years on,'' \emph{Wiley
  Interdisciplinary Reviews: Computational Molecular Science}, vol.~2, no.~1,
  pp. 79--107, 2012.

\bibitem{lavecchia2015machine}
A.~Lavecchia, ``Machine-learning approaches in drug discovery: methods and
  applications,'' \emph{Drug discovery today}, vol.~20, no.~3, pp. 318--331,
  2015.

\bibitem{gawehn2016deep}
E.~Gawehn, J.~A. Hiss, and G.~Schneider, ``Deep learning in drug discovery,''
  \emph{Molecular informatics}, vol.~35, no.~1, pp. 3--14, 2016.

\bibitem{gupta2021artificial}
R.~Gupta, D.~Srivastava, M.~Sahu, S.~Tiwari, R.~K. Ambasta, and P.~Kumar,
  ``Artificial intelligence to deep learning: machine intelligence approach for
  drug discovery,'' \emph{Molecular Diversity}, vol.~25, no.~3, pp. 1315--1360,
  2021.

\bibitem{nigam2021assigning}
A.~Nigam, R.~Pollice, M.~F. Hurley, R.~J. Hickman, M.~Aldeghi, N.~Yoshikawa,
  S.~Chithrananda, V.~A. Voelz, and A.~Aspuru-Guzik, ``Assigning confidence to
  molecular property prediction,'' \emph{Expert opinion on drug discovery},
  vol.~16, no.~9, pp. 1009--1023, 2021.

\bibitem{wieder2020compact}
O.~Wieder, S.~Kohlbacher, M.~Kuenemann, A.~Garon, P.~Ducrot, T.~Seidel, and
  T.~Langer, ``A compact review of molecular property prediction with graph
  neural networks,'' \emph{Drug Discovery Today: Technologies}, vol.~37, pp.
  1--12, 2020.

\bibitem{tong2021generative}
X.~Tong, X.~Liu, X.~Tan, X.~Li, J.~Jiang, Z.~Xiong, T.~Xu, H.~Jiang, N.~Qiao,
  and M.~Zheng, ``Generative models for de novo drug design,'' \emph{Journal of
  Medicinal Chemistry}, vol.~64, no.~19, pp. 14\,011--14\,027, 2021.

\bibitem{elton2019deep}
D.~C. Elton, Z.~Boukouvalas, M.~D. Fuge, and P.~W. Chung, ``Deep learning for
  molecular design—a review of the state of the art,'' \emph{Molecular
  Systems Design \& Engineering}, vol.~4, no.~4, pp. 828--849, 2019.

\bibitem{struble2020current}
T.~J. Struble, J.~C. Alvarez, S.~P. Brown, M.~Chytil, J.~Cisar, R.~L.
  DesJarlais, O.~Engkvist, S.~A. Frank, D.~R. Greve, D.~J. Griffin
  \emph{et~al.}, ``Current and future roles of artificial intelligence in
  medicinal chemistry synthesis,'' \emph{Journal of medicinal chemistry},
  vol.~63, no.~16, pp. 8667--8682, 2020.

\bibitem{dobbelaere2021machine}
M.~R. Dobbelaere, P.~P. Plehiers, R.~Van~de Vijver, C.~V. Stevens, and K.~M.
  Van~Geem, ``Machine learning in chemical engineering: strengths, weaknesses,
  opportunities, and threats,'' \emph{Engineering}, vol.~7, no.~9, pp.
  1201--1211, 2021.

\bibitem{durant2002reoptimization}
J.~L. Durant, B.~A. Leland, D.~R. Henry, and J.~G. Nourse, ``Reoptimization of
  mdl keys for use in drug discovery,'' \emph{Journal of chemical information
  and computer sciences}, vol.~42, no.~6, pp. 1273--1280, 2002.

\bibitem{pubchem}
``Pubchem substructure fingerprint,''
  \url{https://ftp.ncbi.nlm.nih.gov/pubchem/specifications/pubchem_fingerprints.pdf},
  accessed September 2022.

\bibitem{RDKit}
``Rdkit,'' \url{https://www.rdkit.org/}, accessed September 2022.

\bibitem{o2011open}
N.~M. O'Boyle, M.~Banck, C.~A. James, C.~Morley, T.~Vandermeersch, and G.~R.
  Hutchison, ``Open babel: An open chemical toolbox,'' \emph{Journal of
  cheminformatics}, vol.~3, no.~1, pp. 1--14, 2011.

\bibitem{CDK}
``Chemistry development kit (cdk),'' \url{https://cdk.github.io/}, accessed
  September 2022.

\bibitem{xu2017seq2seq}
Z.~Xu, S.~Wang, F.~Zhu, and J.~Huang, ``Seq2seq fingerprint: An unsupervised
  deep molecular embedding for drug discovery,'' in \emph{Proceedings of the
  8th ACM international conference on bioinformatics, computational biology,
  and health informatics}, 2017, pp. 285--294.

\bibitem{jin2018junction}
W.~Jin, R.~Barzilay, and T.~Jaakkola, ``Junction tree variational autoencoder
  for molecular graph generation,'' in \emph{ICML}.\hskip 1em plus 0.5em minus
  0.4em\relax PMLR, 2018, pp. 2323--2332.

\bibitem{weininger1988smiles}
D.~Weininger, ``Smiles, a chemical language and information system. 1.
  introduction to methodology and encoding rules,'' \emph{Journal of chemical
  information and computer sciences}, vol.~28, no.~1, pp. 31--36, 1988.

\bibitem{honda2019smiles}
S.~Honda, S.~Shi, and H.~R. Ueda, ``Smiles transformer: Pre-trained molecular
  fingerprint for low data drug discovery,'' \emph{arXiv preprint
  arXiv:1911.04738}, 2019.

\bibitem{adilov2021generative}
S.~Adilov, ``Generative pre-training from molecules,'' 2021.

\bibitem{radford2019language}
A.~Radford, J.~Wu, R.~Child, D.~Luan, D.~Amodei, I.~Sutskever \emph{et~al.},
  ``Language models are unsupervised multitask learners,'' \emph{OpenAI blog},
  vol.~1, no.~8, p.~9, 2019.

\bibitem{chen2019utilizing}
P.~Chen, W.~Liu, C.-Y. Hsieh, G.~Chen, and S.~Zhang, ``Utilizing edge features
  in graph neural networks via variational information maximization,''
  \emph{arXiv preprint arXiv:1906.05488}, 2019.

\bibitem{olivecrona2017molecular}
M.~Olivecrona, T.~Blaschke, O.~Engkvist, and H.~Chen, ``Molecular de-novo
  design through deep reinforcement learning,'' \emph{Journal of
  cheminformatics}, vol.~9, no.~1, pp. 1--14, 2017.

\bibitem{segler2018generating}
M.~H. Segler, T.~Kogej, C.~Tyrchan, and M.~P. Waller, ``Generating focused
  molecule libraries for drug discovery with recurrent neural networks,''
  \emph{ACS central science}, vol.~4, no.~1, pp. 120--131, 2018.

\bibitem{gupta2018generative}
A.~Gupta, A.~T. M{\"u}ller, B.~J. Huisman, J.~A. Fuchs, P.~Schneider, and
  G.~Schneider, ``Generative recurrent networks for de novo drug design,''
  \emph{Molecular informatics}, vol.~37, no. 1-2, p. 1700111, 2018.

\bibitem{grisoni2020bidirectional}
F.~Grisoni, M.~Moret, R.~Lingwood, and G.~Schneider, ``Bidirectional molecule
  generation with recurrent neural networks,'' \emph{Journal of chemical
  information and modeling}, vol.~60, no.~3, pp. 1175--1183, 2020.

\bibitem{bjerrum2017molecular}
E.~J. Bjerrum and R.~Threlfall, ``Molecular generation with recurrent neural
  networks (rnns),'' \emph{arXiv preprint arXiv:1705.04612}, 2017.

\bibitem{ertl2017silico}
P.~Ertl, R.~Lewis, E.~Martin, and V.~Polyakov, ``In silico generation of novel,
  drug-like chemical matter using the lstm neural network,'' \emph{arXiv
  preprint arXiv:1712.07449}, 2017.

\bibitem{schiff2022augmenting}
Y.~Schiff, V.~Chenthamarakshan, S.~C. Hoffman, K.~N. Ramamurthy, and P.~Das,
  ``Augmenting molecular deep generative models with topological data analysis
  representations,'' in \emph{ICASSP 2022-2022 IEEE International Conference on
  Acoustics, Speech and Signal Processing (ICASSP)}.\hskip 1em plus 0.5em minus
  0.4em\relax IEEE, 2022, pp. 3783--3787.

\bibitem{alperstein2019all}
Z.~Alperstein, A.~Cherkasov, and J.~T. Rolfe, ``All smiles variational
  autoencoder,'' \emph{arXiv preprint arXiv:1905.13343}, 2019.

\bibitem{nigam2019augmenting}
A.~Nigam, P.~Friederich, M.~Krenn, and A.~Aspuru-Guzik, ``Augmenting genetic
  algorithms with deep neural networks for exploring the chemical space,''
  \emph{arXiv preprint arXiv:1909.11655}, 2019.

\bibitem{moss2020boss}
H.~Moss, D.~Leslie, D.~Beck, J.~Gonzalez, and P.~Rayson, ``Boss: Bayesian
  optimization over string spaces,'' \emph{Advances in neural information
  processing systems}, vol.~33, pp. 15\,476--15\,486, 2020.

\bibitem{jin2018learning}
W.~Jin, K.~Yang, R.~Barzilay, and T.~Jaakkola, ``Learning multimodal
  graph-to-graph translation for molecular optimization,'' \emph{arXiv preprint
  arXiv:1812.01070}, 2018.

\bibitem{jin2020hierarchical}
W.~Jin, R.~Barzilay, and T.~Jaakkola, ``Hierarchical generation of molecular
  graphs using structural motifs,'' in \emph{ICML}.\hskip 1em plus 0.5em minus
  0.4em\relax PMLR, 2020, pp. 4839--4848.

\bibitem{maziarz2021learning}
K.~Maziarz, H.~Jackson-Flux, P.~Cameron, F.~Sirockin, N.~Schneider, N.~Stiefl,
  M.~Segler, and M.~Brockschmidt, ``Learning to extend molecular scaffolds with
  structural motifs,'' \emph{arXiv preprint arXiv:2103.03864}, 2021.

\bibitem{you2018graph}
J.~You, B.~Liu, Z.~Ying, V.~Pande, and J.~Leskovec, ``Graph convolutional
  policy network for goal-directed molecular graph generation,'' \emph{Advances
  in neural information processing systems}, vol.~31, 2018.

\bibitem{yang2021hit}
S.~Yang, D.~Hwang, S.~Lee, S.~Ryu, and S.~J. Hwang, ``Hit and lead discovery
  with explorative rl and fragment-based molecule generation,'' \emph{Advances
  in Neural Information Processing Systems}, vol.~34, 2021.

\bibitem{jin2020multi}
W.~Jin, R.~Barzilay, and T.~Jaakkola, ``Multi-objective molecule generation
  using interpretable substructures,'' in \emph{ICML}.\hskip 1em plus 0.5em
  minus 0.4em\relax PMLR, 2020, pp. 4849--4859.

\bibitem{chen2021molecule}
B.~Chen, T.~Wang, C.~Li, H.~Dai, and L.~Song, ``Molecule optimization by
  explainable evolution,'' in \emph{ICLR}, 2021.

\bibitem{xie2021mars}
Y.~Xie, C.~Shi, H.~Zhou, Y.~Yang, W.~Zhang, Y.~Yu, and L.~Li, ``Mars: Markov
  molecular sampling for multi-objective drug discovery,'' \emph{arXiv preprint
  arXiv:2103.10432}, 2021.

\bibitem{andrieu2003introduction}
C.~Andrieu, N.~De~Freitas, A.~Doucet, and M.~I. Jordan, ``An introduction to
  mcmc for machine learning,'' \emph{Machine learning}, vol.~50, no.~1, pp.
  5--43, 2003.

\bibitem{fu2021differentiable}
T.~Fu, W.~Gao, C.~Xiao, J.~Yasonik, C.~W. Coley, and J.~Sun, ``Differentiable
  scaffolding tree for molecular optimization,'' \emph{arXiv preprint
  arXiv:2109.10469}, 2021.

\bibitem{zhou2019optimization}
Z.~Zhou, S.~Kearnes, L.~Li, R.~N. Zare, and P.~Riley, ``Optimization of
  molecules via deep reinforcement learning,'' \emph{Scientific reports},
  vol.~9, no.~1, pp. 1--10, 2019.

\bibitem{shi2020graphaf}
C.~Shi, M.~Xu, Z.~Zhu, W.~Zhang, M.~Zhang, and J.~Tang, ``Graphaf: a flow-based
  autoregressive model for molecular graph generation,'' \emph{arXiv preprint
  arXiv:2001.09382}, 2020.

\bibitem{korovina2020chembo}
K.~Korovina, S.~Xu, K.~Kandasamy, W.~Neiswanger, B.~Poczos, J.~Schneider, and
  E.~Xing, ``Chembo: Bayesian optimization of small organic molecules with
  synthesizable recommendations,'' in \emph{International Conference on
  Artificial Intelligence and Statistics}.\hskip 1em plus 0.5em minus
  0.4em\relax PMLR, 2020, pp. 3393--3403.

\bibitem{chen2021cells}
Z.~Chen, X.~Fang, F.~Wang, X.~Fan, H.~Wu, and H.~Wang, ``Cells: Cost-effective
  evolution in latent space for goal-directed molecular generation,''
  \emph{arXiv preprint arXiv:2112.00905}, 2021.

\bibitem{wu2021spatial}
Y.~Wu, N.~Choma, A.~Chen, M.~Cashman, {\'E}.~T. Prates, M.~Shah, V.~G.~M.
  Vergara, A.~Clyde, T.~S. Brettin, W.~A. de~Jong \emph{et~al.}, ``Spatial
  graph attention and curiosity-driven policy for antiviral drug discovery,''
  \emph{arXiv preprint arXiv:2106.02190}, 2021.

\bibitem{simm2019generative}
G.~N. Simm and J.~M. Hern{\'a}ndez-Lobato, ``A generative model for molecular
  distance geometry,'' \emph{arXiv preprint arXiv:1909.11459}, 2019.

\bibitem{xu2021end}
M.~Xu, W.~Wang, S.~Luo, C.~Shi, Y.~Bengio, R.~Gomez-Bombarelli, and J.~Tang,
  ``An end-to-end framework for molecular conformation generation via bilevel
  programming,'' in \emph{ICML}.\hskip 1em plus 0.5em minus 0.4em\relax PMLR,
  2021, pp. 11\,537--11\,547.

\bibitem{shi2021learning}
C.~Shi, S.~Luo, M.~Xu, and J.~Tang, ``Learning gradient fields for molecular
  conformation generation,'' in \emph{ICML}.\hskip 1em plus 0.5em minus
  0.4em\relax PMLR, 2021, pp. 9558--9568.

\bibitem{zhu2022direct}
J.~Zhu, Y.~Xia, C.~Liu, L.~Wu, S.~Xie, T.~Wang, Y.~Wang, W.~Zhou, T.~Qin, H.~Li
  \emph{et~al.}, ``Direct molecular conformation generation,'' \emph{arXiv
  preprint arXiv:2202.01356}, 2022.

\bibitem{xu2022geodiff}
M.~Xu, L.~Yu, Y.~Song, C.~Shi, S.~Ermon, and J.~Tang, ``Geodiff: A geometric
  diffusion model for molecular conformation generation,'' \emph{arXiv preprint
  arXiv:2203.02923}, 2022.

\bibitem{luo2021autoregressive}
Y.~Luo and S.~Ji, ``An autoregressive flow model for 3d molecular geometry
  generation from scratch,'' in \emph{ICLR}, 2021.

\bibitem{mansimov2019molecular}
E.~Mansimov, O.~Mahmood, S.~Kang, and K.~Cho, ``Molecular geometry prediction
  using a deep generative graph neural network,'' \emph{Scientific reports},
  vol.~9, no.~1, pp. 1--13, 2019.

\bibitem{guan2021energy}
J.~Guan, W.~W. Qian, W.-Y. Ma, J.~Ma, J.~Peng \emph{et~al.}, ``Energy-inspired
  molecular conformation optimization,'' in \emph{ICLR}, 2021.

\bibitem{xu2021learning}
M.~Xu, S.~Luo, Y.~Bengio, J.~Peng, and J.~Tang, ``Learning neural generative
  dynamics for molecular conformation generation,'' \emph{arXiv preprint
  arXiv:2102.10240}, 2021.

\bibitem{hochreiter1997long}
S.~Hochreiter and J.~Schmidhuber, ``Long short-term memory,'' \emph{Neural
  computation}, vol.~9, no.~8, pp. 1735--1780, 1997.

\bibitem{cho2014properties}
K.~Cho, B.~Van~Merri{\"e}nboer, D.~Bahdanau, and Y.~Bengio, ``On the properties
  of neural machine translation: Encoder-decoder approaches,'' \emph{arXiv
  preprint arXiv:1409.1259}, 2014.

\bibitem{mayr2018large}
A.~Mayr, G.~Klambauer, T.~Unterthiner, M.~Steijaert, J.~K. Wegner,
  H.~Ceulemans, D.-A. Clevert, and S.~Hochreiter, ``Large-scale comparison of
  machine learning methods for drug target prediction on chembl,''
  \emph{Chemical science}, vol.~9, no.~24, pp. 5441--5451, 2018.

\bibitem{goh2017smiles2vec}
G.~B. Goh, N.~O. Hodas, C.~Siegel, and A.~Vishnu, ``Smiles2vec: An
  interpretable general-purpose deep neural network for predicting chemical
  properties,'' \emph{arXiv preprint arXiv:1712.02034}, 2017.

\bibitem{li2022novel}
C.~Li, J.~Feng, S.~Liu, and J.~Yao, ``A novel molecular representation learning
  for molecular property prediction with a multiple smiles-based
  augmentation,'' \emph{Computational Intelligence and Neuroscience}, vol.
  2022, 2022.

\bibitem{peng2019top}
Y.~Peng, Z.~Zhang, Q.~Jiang, J.~Guan, and S.~Zhou, ``Top: Towards better
  toxicity prediction by deep molecular representation learning,'' in
  \emph{2019 IEEE International Conference on Bioinformatics and
  Biomedicine}.\hskip 1em plus 0.5em minus 0.4em\relax IEEE, 2019, pp.
  318--325.

\bibitem{paul2018chemixnet}
A.~Paul, D.~Jha, R.~Al-Bahrani, W.-k. Liao, A.~Choudhary, and A.~Agrawal,
  ``Chemixnet: Mixed dnn architectures for predicting chemical properties using
  multiple molecular representations,'' \emph{arXiv preprint arXiv:1811.08283},
  2018.

\bibitem{goh2017chemception}
G.~B. Goh, C.~Siegel, A.~Vishnu, N.~O. Hodas, and N.~Baker, ``Chemception: a
  deep neural network with minimal chemistry knowledge matches the performance
  of expert-developed qsar/qspr models,'' \emph{arXiv preprint
  arXiv:1706.06689}, 2017.

\bibitem{kingma2014semi}
D.~P. Kingma, S.~Mohamed, D.~Jimenez~Rezende, and M.~Welling, ``Semi-supervised
  learning with deep generative models,'' \emph{Advances in neural information
  processing systems}, vol.~27, 2014.

\bibitem{devlin2018bert}
J.~Devlin, M.-W. Chang, K.~Lee, and K.~Toutanova, ``Bert: Pre-training of deep
  bidirectional transformers for language understanding,'' \emph{arXiv preprint
  arXiv:1810.04805}, 2018.

\bibitem{liu2019roberta}
Y.~Liu, M.~Ott, N.~Goyal, J.~Du, M.~Joshi, D.~Chen, O.~Levy, M.~Lewis,
  L.~Zettlemoyer, and V.~Stoyanov, ``Roberta: A robustly optimized bert
  pretraining approach,'' \emph{arXiv preprint arXiv:1907.11692}, 2019.

\bibitem{rahimovich2021predicting}
D.~R. Rahimovich, S.~R. Abdiqayum~o'g, A.~S. Qaxramon~o'g'li \emph{et~al.},
  ``Predicting the activity and properties of chemicals based on roberta,'' in
  \emph{2021 International Conference on Information Science and Communications
  Technologies (ICISCT)}.\hskip 1em plus 0.5em minus 0.4em\relax IEEE, 2021,
  pp. 1--4.

\bibitem{chithrananda2020chemberta}
S.~Chithrananda, G.~Grand, and B.~Ramsundar, ``Chemberta: large-scale
  self-supervised pretraining for molecular property prediction,'' \emph{arXiv
  preprint arXiv:2010.09885}, 2020.

\bibitem{unterthiner2014deep}
T.~Unterthiner, A.~Mayr, G.~Klambauer, M.~Steijaert, J.~K. Wegner,
  H.~Ceulemans, and S.~Hochreiter, ``Deep learning as an opportunity in virtual
  screening,'' in \emph{Proceedings of the deep learning workshop at NIPS},
  vol.~27, 2014, pp. 1--9.

\bibitem{mayr2016deeptox}
A.~Mayr, G.~Klambauer, T.~Unterthiner, and S.~Hochreiter, ``Deeptox: toxicity
  prediction using deep learning,'' \emph{Frontiers in Environmental Science},
  vol.~3, p.~80, 2016.

\bibitem{schimunek2022context}
J.~Schimunek, P.~Seidl, L.~Friedrich, D.~Kuhn, F.~Rippmann, S.~Hochreiter, and
  G.~Klambauer, ``Context-enriched molecule representations improve few-shot
  drug discovery,'' 2022.

\bibitem{zhang2018seq3seq}
X.~Zhang, S.~Wang, F.~Zhu, Z.~Xu, Y.~Wang, and J.~Huang, ``Seq3seq fingerprint:
  towards end-to-end semi-supervised deep drug discovery,'' in
  \emph{Proceedings of the 2018 ACM International Conference on Bioinformatics,
  Computational Biology, and Health Informatics}, 2018, pp. 404--413.

\bibitem{wang2019smiles}
S.~Wang, Y.~Guo, Y.~Wang, H.~Sun, and J.~Huang, ``Smiles-bert: large scale
  unsupervised pre-training for molecular property prediction,'' in
  \emph{Proceedings of the 10th ACM international conference on bioinformatics,
  computational biology and health informatics}, 2019, pp. 429--436.

\bibitem{glen2006circular}
R.~C. Glen, A.~Bender, C.~H. Arnby, L.~Carlsson, S.~Boyer, and J.~Smith,
  ``Circular fingerprints: flexible molecular descriptors with applications
  from physical chemistry to adme,'' \emph{IDrugs}, vol.~9, no.~3, p. 199,
  2006.

\bibitem{bahdanau2014neural}
D.~Bahdanau, K.~Cho, and Y.~Bengio, ``Neural machine translation by jointly
  learning to align and translate,'' \emph{arXiv preprint arXiv:1409.0473},
  2014.

\bibitem{kipf2016semi}
T.~N. Kipf and M.~Welling, ``Semi-supervised classification with graph
  convolutional networks,'' \emph{arXiv preprint arXiv:1609.02907}, 2016.

\bibitem{ramakrishnan2014quantum}
R.~Ramakrishnan, P.~O. Dral, M.~Rupp, and O.~A. Von~Lilienfeld, ``Quantum
  chemistry structures and properties of 134 kilo molecules,'' \emph{Scientific
  data}, vol.~1, no.~1, pp. 1--7, 2014.

\bibitem{lu2019molecular}
C.~Lu, Q.~Liu, C.~Wang, Z.~Huang, P.~Lin, and L.~He, ``Molecular property
  prediction: A multilevel quantum interactions modeling perspective,'' in
  \emph{Proceedings of the AAAI Conference on Artificial Intelligence},
  vol.~33, no.~01, 2019, pp. 1052--1060.

\bibitem{liu2021pre}
S.~Liu, H.~Wang, W.~Liu, J.~Lasenby, H.~Guo, and J.~Tang, ``Pre-training
  molecular graph representation with 3d geometry,'' \emph{arXiv preprint
  arXiv:2110.07728}, 2021.

\bibitem{hu2019strategies}
W.~Hu, B.~Liu, J.~Gomes, M.~Zitnik, P.~Liang, V.~Pande, and J.~Leskovec,
  ``Strategies for pre-training graph neural networks,'' \emph{arXiv preprint
  arXiv:1905.12265}, 2019.

\bibitem{zeng2022accurate}
X.~Zeng, H.~Xiang, L.~Yu, J.~Wang, K.~Li, R.~Nussinov, and F.~Cheng, ``Accurate
  prediction of molecular properties and drug targets using a self-supervised
  image representation learning framework,'' \emph{Nature Machine
  Intelligence}, pp. 1--13, 2022.

\bibitem{rong2020self}
Y.~Rong, Y.~Bian, T.~Xu, W.~Xie, Y.~Wei, W.~Huang, and J.~Huang,
  ``Self-supervised graph transformer on large-scale molecular data,''
  \emph{Advances in Neural Information Processing Systems}, vol.~33, pp.
  12\,559--12\,571, 2020.

\bibitem{you2020graph}
Y.~You, T.~Chen, Y.~Sui, T.~Chen, Z.~Wang, and Y.~Shen, ``Graph contrastive
  learning with augmentations,'' \emph{Advances in Neural Information
  Processing Systems}, vol.~33, pp. 5812--5823, 2020.

\bibitem{wang2022molecular}
Y.~Wang, J.~Wang, Z.~Cao, and A.~Barati~Farimani, ``Molecular contrastive
  learning of representations via graph neural networks,'' \emph{Nature Machine
  Intelligence}, vol.~4, no.~3, pp. 279--287, 2022.

\bibitem{li2021geomgcl}
S.~Li, J.~Zhou, T.~Xu, D.~Dou, and H.~Xiong, ``Geomgcl: Geometric graph
  contrastive learning for molecular property prediction,'' \emph{arXiv
  preprint arXiv:2109.11730}, 2021.

\bibitem{liu2019n}
S.~Liu, M.~F. Demirel, and Y.~Liang, ``N-gram graph: Simple unsupervised
  representation for graphs, with applications to molecules,'' \emph{Advances
  in neural information processing systems}, vol.~32, 2019.

\bibitem{altae2017low}
H.~Altae-Tran, B.~Ramsundar, A.~S. Pappu, and V.~Pande, ``Low data drug
  discovery with one-shot learning,'' \emph{ACS central science}, vol.~3,
  no.~4, pp. 283--293, 2017.

\bibitem{gilmer2017neural}
J.~Gilmer, S.~S. Schoenholz, P.~F. Riley, O.~Vinyals, and G.~E. Dahl, ``Neural
  message passing for quantum chemistry,'' in \emph{ICML}.\hskip 1em plus 0.5em
  minus 0.4em\relax PMLR, 2017, pp. 1263--1272.

\bibitem{liu2021spherical}
Y.~Liu, L.~Wang, M.~Liu, Y.~Lin, X.~Zhang, B.~Oztekin, and S.~Ji, ``Spherical
  message passing for 3d molecular graphs,'' in \emph{ICLR}, 2021.

\bibitem{hao2020asgn}
Z.~Hao, C.~Lu, Z.~Huang, H.~Wang, Z.~Hu, Q.~Liu, E.~Chen, and C.~Lee, ``Asgn:
  An active semi-supervised graph neural network for molecular property
  prediction,'' in \emph{Proceedings of the 26th ACM SIGKDD International
  Conference on Knowledge Discovery \& Data Mining}, 2020, pp. 731--752.

\bibitem{dwivedi2021graph}
V.~P. Dwivedi, A.~T. Luu, T.~Laurent, Y.~Bengio, and X.~Bresson, ``Graph neural
  networks with learnable structural and positional representations,''
  \emph{arXiv preprint arXiv:2110.07875}, 2021.

\bibitem{velickovic2019deep}
P.~Velickovic, W.~Fedus, W.~L. Hamilton, P.~Li{\`o}, Y.~Bengio, and R.~D.
  Hjelm, ``Deep graph infomax.'' \emph{ICLR (Poster)}, vol.~2, no.~3, p.~4,
  2019.

\bibitem{hjelm2018learning}
R.~D. Hjelm, A.~Fedorov, S.~Lavoie-Marchildon, K.~Grewal, P.~Bachman,
  A.~Trischler, and Y.~Bengio, ``Learning deep representations by mutual
  information estimation and maximization,'' \emph{arXiv preprint
  arXiv:1808.06670}, 2018.

\bibitem{xu2018powerful}
K.~Xu, W.~Hu, J.~Leskovec, and S.~Jegelka, ``How powerful are graph neural
  networks?'' \emph{arXiv preprint arXiv:1810.00826}, 2018.

\bibitem{sun2019infograph}
F.-Y. Sun, J.~Hoffmann, V.~Verma, and J.~Tang, ``Infograph: Unsupervised and
  semi-supervised graph-level representation learning via mutual information
  maximization,'' \emph{arXiv preprint arXiv:1908.01000}, 2019.

\bibitem{li2020conformation}
Z.~Li, S.~Yang, G.~Song, and L.~Cai, ``Conformation-guided molecular
  representation with hamiltonian neural networks,'' in \emph{ICLR}, 2020.

\bibitem{yang2021deep}
S.~Yang, Z.~Li, G.~Song, and L.~Cai, ``Deep molecular representation learning
  via fusing physical and chemical information,'' \emph{Advances in Neural
  Information Processing Systems}, vol.~34, 2021.

\bibitem{maziarka2020molecule}
{\L}.~Maziarka, T.~Danel, S.~Mucha, K.~Rataj, J.~Tabor, and
  S.~Jastrz{\k{e}}bski, ``Molecule attention transformer,'' \emph{arXiv
  preprint arXiv:2002.08264}, 2020.

\bibitem{segler2017neural}
M.~H. Segler and M.~P. Waller, ``Neural-symbolic machine learning for
  retrosynthesis and reaction prediction,'' \emph{Chemistry--A European
  Journal}, vol.~23, no.~25, pp. 5966--5971, 2017.

\bibitem{coley2017computer}
C.~W. Coley, L.~Rogers, W.~H. Green, and K.~F. Jensen, ``Computer-assisted
  retrosynthesis based on molecular similarity,'' \emph{ACS central science},
  vol.~3, no.~12, pp. 1237--1245, 2017.

\bibitem{baylon2019enhancing}
J.~L. Baylon, N.~A. Cilfone, J.~R. Gulcher, and T.~W. Chittenden, ``Enhancing
  retrosynthetic reaction prediction with deep learning using multiscale
  reaction classification,'' \emph{Journal of chemical information and
  modeling}, vol.~59, no.~2, pp. 673--688, 2019.

\bibitem{segler2018planning}
M.~H. Segler, M.~Preuss, and M.~P. Waller, ``Planning chemical syntheses with
  deep neural networks and symbolic ai,'' \emph{Nature}, vol. 555, no. 7698,
  pp. 604--610, 2018.

\bibitem{seidl2022improving}
P.~Seidl, P.~Renz, N.~Dyubankova, P.~Neves, J.~Verhoeven, J.~K. Wegner,
  M.~Segler, S.~Hochreiter, and G.~Klambauer, ``Improving few-and zero-shot
  reaction template prediction using modern hopfield networks,'' \emph{Journal
  of chemical information and modeling}, vol.~62, no.~9, pp. 2111--2120, 2022.

\bibitem{dai2019retrosynthesis}
H.~Dai, C.~Li, C.~Coley, B.~Dai, and L.~Song, ``Retrosynthesis prediction with
  conditional graph logic network,'' \emph{Advances in Neural Information
  Processing Systems}, vol.~32, 2019.

\bibitem{chen2021deep}
S.~Chen and Y.~Jung, ``Deep retrosynthetic reaction prediction using local
  reactivity and global attention,'' \emph{JACS Au}, vol.~1, no.~10, pp.
  1612--1620, 2021.

\bibitem{liu2017retrosynthetic}
B.~Liu, B.~Ramsundar, P.~Kawthekar, J.~Shi, J.~Gomes, Q.~Luu~Nguyen, S.~Ho,
  J.~Sloane, P.~Wender, and V.~Pande, ``Retrosynthetic reaction prediction
  using neural sequence-to-sequence models,'' \emph{ACS central science},
  vol.~3, no.~10, pp. 1103--1113, 2017.

\bibitem{karpov2019transformer}
P.~Karpov, G.~Godin, and I.~V. Tetko, ``A transformer model for
  retrosynthesis,'' in \emph{International Conference on Artificial Neural
  Networks}.\hskip 1em plus 0.5em minus 0.4em\relax Springer, 2019, pp.
  817--830.

\bibitem{chen2019learning}
B.~Chen, T.~Shen, T.~S. Jaakkola, and R.~Barzilay, ``Learning to make
  generalizable and diverse predictions for retrosynthesis,'' \emph{arXiv
  preprint arXiv:1910.09688}, 2019.

\bibitem{ishiguro2020data}
K.~Ishiguro, K.~Ujihara, R.~Sawada, H.~Akita, and M.~Kotera, ``Data transfer
  approaches to improve seq-to-seq retrosynthesis,'' \emph{arXiv preprint
  arXiv:2010.00792}, 2020.

\bibitem{zheng2019predicting}
S.~Zheng, J.~Rao, Z.~Zhang, J.~Xu, and Y.~Yang, ``Predicting retrosynthetic
  reactions using self-corrected transformer neural networks,'' \emph{Journal
  of chemical information and modeling}, vol.~60, no.~1, pp. 47--55, 2019.

\bibitem{zhao2022leveraging}
M.~Zhao, L.~Fang, L.~Tan, J.-G. Lou, and Y.~Lepage, ``Leveraging reaction-aware
  substructures for retrosynthesis and reaction prediction,'' \emph{arXiv
  preprint arXiv:2204.05919}, 2022.

\bibitem{sacha2021molecule}
M.~Sacha, M.~B{\l}az, P.~Byrski, P.~Dabrowski-Tumanski, M.~Chrominski,
  R.~Loska, P.~W{\l}odarczyk-Pruszynski, and S.~Jastrzebski, ``Molecule edit
  graph attention network: modeling chemical reactions as sequences of graph
  edits,'' \emph{Journal of Chemical Information and Modeling}, vol.~61, no.~7,
  pp. 3273--3284, 2021.

\bibitem{shi2020graph}
C.~Shi, M.~Xu, H.~Guo, M.~Zhang, and J.~Tang, ``A graph to graphs framework for
  retrosynthesis prediction,'' in \emph{ICML}.\hskip 1em plus 0.5em minus
  0.4em\relax PMLR, 2020, pp. 8818--8827.

\bibitem{somnath2021learning}
V.~R. Somnath, C.~Bunne, C.~Coley, A.~Krause, and R.~Barzilay, ``Learning graph
  models for retrosynthesis prediction,'' \emph{Advances in Neural Information
  Processing Systems}, vol.~34, 2021.

\bibitem{gao2022semiretro}
Z.~Gao, C.~Tan, L.~Wu, and S.~Z. Li, ``Semiretro: Semi-template framework
  boosts deep retrosynthesis prediction,'' \emph{arXiv preprint
  arXiv:2202.08205}, 2022.

\bibitem{yan2020retroxpert}
C.~Yan, Q.~Ding, P.~Zhao, S.~Zheng, J.~Yang, Y.~Yu, and J.~Huang, ``Retroxpert:
  Decompose retrosynthesis prediction like a chemist,'' \emph{Advances in
  Neural Information Processing Systems}, vol.~33, pp. 11\,248--11\,258, 2020.

\bibitem{sun2020energy}
R.~Sun, H.~Dai, L.~Li, S.~Kearnes, and B.~Dai, ``Energy-based view of
  retrosynthesis,'' \emph{arXiv preprint arXiv:2007.13437}, 2020.

\bibitem{seo2021gta}
S.-W. Seo, Y.~Y. Song, J.~Y. Yang, S.~Bae, H.~Lee, J.~Shin, S.~J. Hwang, and
  E.~Yang, ``Gta: Graph truncated attention for retrosynthesis,'' in
  \emph{Proceedings of the AAAI Conference on Artificial Intelligence},
  vol.~35, no.~1, 2021, pp. 531--539.

\bibitem{wan2022retroformer}
Y.~Wan, B.~Liao, C.-Y. Hsieh, and S.~Zhang, ``Retroformer: Pushing the limits
  of interpretable end-to-end retrosynthesis transformer,'' \emph{arXiv
  preprint arXiv:2201.12475}, 2022.

\bibitem{wei2016neural}
J.~N. Wei, D.~Duvenaud, and A.~Aspuru-Guzik, ``Neural networks for the
  prediction of organic chemistry reactions,'' \emph{ACS central science},
  vol.~2, no.~10, pp. 725--732, 2016.

\bibitem{coley2017prediction}
C.~W. Coley, R.~Barzilay, T.~S. Jaakkola, W.~H. Green, and K.~F. Jensen,
  ``Prediction of organic reaction outcomes using machine learning,'' \emph{ACS
  central science}, vol.~3, no.~5, pp. 434--443, 2017.

\bibitem{segler2017modelling}
M.~H. Segler and M.~P. Waller, ``Modelling chemical reasoning to predict and
  invent reactions,'' \emph{Chemistry--A European Journal}, vol.~23, no.~25,
  pp. 6118--6128, 2017.

\bibitem{jin2017predicting}
W.~Jin, C.~Coley, R.~Barzilay, and T.~Jaakkola, ``Predicting organic reaction
  outcomes with weisfeiler-lehman network,'' \emph{Advances in neural
  information processing systems}, vol.~30, 2017.

\bibitem{coley2019graph}
C.~W. Coley, W.~Jin, L.~Rogers, T.~F. Jamison, T.~S. Jaakkola, W.~H. Green,
  R.~Barzilay, and K.~F. Jensen, ``A graph-convolutional neural network model
  for the prediction of chemical reactivity,'' \emph{Chemical science},
  vol.~10, no.~2, pp. 370--377, 2019.

\bibitem{qian2020integrating}
W.~W. Qian, N.~T. Russell, C.~L. Simons, Y.~Luo, M.~D. Burke, and J.~Peng,
  ``Integrating deep neural networks and symbolic inference for organic
  reactivity prediction,'' 2020.

\bibitem{bradshaw2018generative}
J.~Bradshaw, M.~J. Kusner, B.~Paige, M.~H. Segler, and J.~M.
  Hern{\'a}ndez-Lobato, ``A generative model for electron paths,'' \emph{arXiv
  preprint arXiv:1805.10970}, 2018.

\bibitem{do2019graph}
K.~Do, T.~Tran, and S.~Venkatesh, ``Graph transformation policy network for
  chemical reaction prediction,'' in \emph{Proceedings of the 25th ACM SIGKDD
  International Conference on Knowledge Discovery \& Data Mining}, 2019, pp.
  750--760.

\bibitem{bi2021non}
H.~Bi, H.~Wang, C.~Shi, C.~Coley, J.~Tang, and H.~Guo, ``Non-autoregressive
  electron redistribution modeling for reaction prediction,'' in
  \emph{ICML}.\hskip 1em plus 0.5em minus 0.4em\relax PMLR, 2021, pp. 904--913.

\bibitem{schwaller2018found}
P.~Schwaller, T.~Gaudin, D.~Lanyi, C.~Bekas, and T.~Laino, ``“found in
  translation”: predicting outcomes of complex organic chemistry reactions
  using neural sequence-to-sequence models,'' \emph{Chemical science}, vol.~9,
  no.~28, pp. 6091--6098, 2018.

\bibitem{kim2021pubchem}
S.~Kim, J.~Chen, T.~Cheng, A.~Gindulyte, J.~He, S.~He, Q.~Li, B.~A. Shoemaker,
  P.~A. Thiessen, B.~Yu \emph{et~al.}, ``Pubchem in 2021: new data content and
  improved web interfaces,'' \emph{Nucleic acids research}, vol.~49, no.~D1,
  pp. D1388--D1395, 2021.

\bibitem{gaulton2012chembl}
A.~Gaulton, L.~J. Bellis, A.~P. Bento, J.~Chambers, M.~Davies, A.~Hersey,
  Y.~Light, S.~McGlinchey, D.~Michalovich, B.~Al-Lazikani \emph{et~al.},
  ``Chembl: a large-scale bioactivity database for drug discovery,''
  \emph{Nucleic acids research}, vol.~40, no.~D1, pp. D1100--D1107, 2012.

\bibitem{sterling2015zinc}
T.~Sterling and J.~J. Irwin, ``Zinc 15--ligand discovery for everyone,''
  \emph{Journal of chemical information and modeling}, vol.~55, no.~11, pp.
  2324--2337, 2015.

\bibitem{fink2005virtual}
T.~Fink, H.~Bruggesser, and J.-L. Reymond, ``Virtual exploration of the
  small-molecule chemical universe below 160 daltons,'' \emph{Angewandte Chemie
  International Edition}, vol.~44, no.~10, pp. 1504--1508, 2005.

\bibitem{fink2007virtual}
T.~Fink and J.-L. Reymond, ``Virtual exploration of the chemical universe up to
  11 atoms of c, n, o, f: assembly of 26.4 million structures (110.9 million
  stereoisomers) and analysis for new ring systems, stereochemistry,
  physicochemical properties, compound classes, and drug discovery,''
  \emph{Journal of chemical information and modeling}, vol.~47, no.~2, pp.
  342--353, 2007.

\bibitem{blum2009970}
L.~C. Blum and J.-L. Reymond, ``970 million druglike small molecules for
  virtual screening in the chemical universe database gdb-13,'' \emph{Journal
  of the American Chemical Society}, vol. 131, no.~25, pp. 8732--8733, 2009.

\bibitem{ruddigkeit2012enumeration}
L.~Ruddigkeit, R.~Van~Deursen, L.~C. Blum, and J.-L. Reymond, ``Enumeration of
  166 billion organic small molecules in the chemical universe database
  gdb-17,'' \emph{Journal of chemical information and modeling}, vol.~52,
  no.~11, pp. 2864--2875, 2012.

\bibitem{axelrod2022geom}
S.~Axelrod and R.~Gomez-Bombarelli, ``Geom, energy-annotated molecular
  conformations for property prediction and molecular generation,''
  \emph{Scientific Data}, vol.~9, no.~1, pp. 1--14, 2022.

\bibitem{schneider2016s}
N.~Schneider, N.~Stiefl, and G.~A. Landrum, ``What’s what: The (nearly)
  definitive guide to reaction role assignment,'' \emph{Journal of chemical
  information and modeling}, vol.~56, no.~12, pp. 2336--2346, 2016.

\bibitem{wu2018moleculenet}
Z.~Wu, B.~Ramsundar, E.~N. Feinberg, J.~Gomes, C.~Geniesse, A.~S. Pappu,
  K.~Leswing, and V.~Pande, ``Moleculenet: a benchmark for molecular machine
  learning,'' \emph{Chemical science}, vol.~9, no.~2, pp. 513--530, 2018.

\bibitem{sun2017excape}
J.~Sun, N.~Jeliazkova, V.~Chupakhin, J.-F. Golib-Dzib, O.~Engkvist,
  L.~Carlsson, J.~Wegner, H.~Ceulemans, I.~Georgiev, V.~Jeliazkov
  \emph{et~al.}, ``Excape-db: an integrated large scale dataset facilitating
  big data analysis in chemogenomics,'' \emph{Journal of cheminformatics},
  vol.~9, pp. 1--9, 2017.

\bibitem{walters2020assessing}
W.~P. Walters and M.~Murcko, ``Assessing the impact of generative ai on
  medicinal chemistry,'' \emph{Nature biotechnology}, vol.~38, no.~2, pp.
  143--145, 2020.

\bibitem{brown2019guacamol}
N.~Brown, M.~Fiscato, M.~H. Segler, and A.~C. Vaucher, ``Guacamol: benchmarking
  models for de novo molecular design,'' \emph{Journal of chemical information
  and modeling}, vol.~59, no.~3, pp. 1096--1108, 2019.

\bibitem{stanley2021fs}
M.~Stanley, J.~F. Bronskill, K.~Maziarz, H.~Misztela, J.~Lanini, M.~Segler,
  N.~Schneider, and M.~Brockschmidt, ``Fs-mol: A few-shot learning dataset of
  molecules,'' in \emph{Thirty-fifth Conference on Neural Information
  Processing Systems Datasets and Benchmarks Track (Round 2)}, 2021.

\bibitem{subramanian2016computational}
G.~Subramanian, B.~Ramsundar, V.~Pande, and R.~A. Denny, ``Computational
  modeling of $\beta$-secretase 1 (bace-1) inhibitors using ligand based
  approaches,'' \emph{Journal of chemical information and modeling}, vol.~56,
  no.~10, pp. 1936--1949, 2016.

\bibitem{guo2022data}
M.~Guo, V.~Thost, B.~Li, P.~Das, J.~Chen, and W.~Matusik, ``Data-efficient
  graph grammar learning for molecular generation,'' \emph{arXiv preprint
  arXiv:2203.08031}, 2022.

\bibitem{chen2020retro}
B.~Chen, C.~Li, H.~Dai, and L.~Song, ``Retro*: learning retrosynthetic planning
  with neural guided a* search,'' in \emph{ICML}.\hskip 1em plus 0.5em minus
  0.4em\relax PMLR, 2020, pp. 1608--1616.

\bibitem{sidorov2019predicting}
P.~Sidorov, S.~Naulaerts, J.~Ariey-Bonnet, E.~Pasquier, and P.~J. Ballester,
  ``Predicting synergism of cancer drug combinations using nci-almanac data,''
  \emph{Frontiers in chemistry}, vol.~7, p. 509, 2019.

\bibitem{safikhani2016revisiting}
Z.~Safikhani, P.~Smirnov, M.~Freeman, N.~El-Hachem, A.~She, Q.~Rene,
  A.~Goldenberg, N.~J. Birkbak, C.~Hatzis, L.~Shi \emph{et~al.}, ``Revisiting
  inconsistency in large pharmacogenomic studies,'' \emph{F1000Research},
  vol.~5, 2016.

\bibitem{renz2019failure}
P.~Renz, D.~Van~Rompaey, J.~K. Wegner, S.~Hochreiter, and G.~Klambauer, ``On
  failure modes in molecule generation and optimization,'' \emph{Drug Discovery
  Today: Technologies}, vol.~32, pp. 55--63, 2019.

\bibitem{schwaller2020predicting}
P.~Schwaller, R.~Petraglia, V.~Zullo, V.~H. Nair, R.~A. Haeuselmann, R.~Pisoni,
  C.~Bekas, A.~Iuliano, and T.~Laino, ``Predicting retrosynthetic pathways
  using transformer-based models and a hyper-graph exploration strategy,''
  \emph{Chemical science}, vol.~11, no.~12, pp. 3316--3325, 2020.

\bibitem{thakkar2020datasets}
A.~Thakkar, T.~Kogej, J.-L. Reymond, O.~Engkvist, and E.~J. Bjerrum, ``Datasets
  and their influence on the development of computer assisted synthesis
  planning tools in the pharmaceutical domain,'' \emph{Chemical science},
  vol.~11, no.~1, pp. 154--168, 2020.

\bibitem{fortunato2020data}
M.~E. Fortunato, C.~W. Coley, B.~C. Barnes, and K.~F. Jensen, ``Data
  augmentation and pretraining for template-based retrosynthetic prediction in
  computer-aided synthesis planning,'' \emph{Journal of chemical information
  and modeling}, vol.~60, no.~7, pp. 3398--3407, 2020.

\bibitem{jimenez2020drug}
J.~Jim{\'e}nez-Luna, F.~Grisoni, and G.~Schneider, ``Drug discovery with
  explainable artificial intelligence,'' \emph{Nature Machine Intelligence},
  vol.~2, no.~10, pp. 573--584, 2020.

\bibitem{vamathevan2019applications}
J.~Vamathevan, D.~Clark, P.~Czodrowski, I.~Dunham, E.~Ferran, G.~Lee, B.~Li,
  A.~Madabhushi, P.~Shah, M.~Spitzer \emph{et~al.}, ``Applications of machine
  learning in drug discovery and development,'' \emph{Nature reviews Drug
  discovery}, vol.~18, no.~6, pp. 463--477, 2019.

\bibitem{preuer2019interpretable}
K.~Preuer, G.~Klambauer, F.~Rippmann, S.~Hochreiter, and T.~Unterthiner,
  ``Interpretable deep learning in drug discovery,'' \emph{Explainable AI:
  interpreting, explaining and visualizing deep learning}, pp. 331--345, 2019.

\bibitem{du2022interpretable}
Y.~Du, X.~Guo, A.~Shehu, and L.~Zhao, ``Interpretable molecular graph
  generation via monotonic constraints,'' in \emph{Proceedings of the 2022 SIAM
  International Conference on Data Mining (SDM)}.\hskip 1em plus 0.5em minus
  0.4em\relax SIAM, 2022, pp. 73--81.

\bibitem{du2022molgensurvey}
Y.~Du, T.~Fu, J.~Sun, and S.~Liu, ``Molgensurvey: A systematic survey in
  machine learning models for molecule design,'' \emph{arXiv preprint
  arXiv:2203.14500}, 2022.

\bibitem{mervin2021uncertainty}
L.~H. Mervin, S.~Johansson, E.~Semenova, K.~A. Giblin, and O.~Engkvist,
  ``Uncertainty quantification in drug design,'' \emph{Drug discovery today},
  vol.~26, no.~2, pp. 474--489, 2021.

\bibitem{hernandez2022conformal}
S.~Hern{\'a}ndez-Hern{\'a}ndez, S.~Vishwakarma, and P.~Ballester, ``Conformal
  prediction of small-molecule drug resistance in cancer cell lines,'' in
  \emph{Conformal and Probabilistic Prediction with Applications}.\hskip 1em
  plus 0.5em minus 0.4em\relax PMLR, 2022, pp. 92--108.

\bibitem{hie2020leveraging}
B.~Hie, B.~D. Bryson, and B.~Berger, ``Leveraging uncertainty in machine
  learning accelerates biological discovery and design,'' \emph{Cell systems},
  vol.~11, no.~5, pp. 461--477, 2020.

\bibitem{vaucher2021inferring}
A.~C. Vaucher, P.~Schwaller, J.~Geluykens, V.~H. Nair, A.~Iuliano, and
  T.~Laino, ``Inferring experimental procedures from text-based representations
  of chemical reactions,'' \emph{Nature communications}, vol.~12, no.~1, pp.
  1--11, 2021.

\bibitem{wang2020towards}
X.~Wang, Y.~Qian, H.~Gao, C.~W. Coley, Y.~Mo, R.~Barzilay, and K.~F. Jensen,
  ``Towards efficient discovery of green synthetic pathways with monte carlo
  tree search and reinforcement learning,'' \emph{Chemical science}, vol.~11,
  no.~40, pp. 10\,959--10\,972, 2020.

\bibitem{henson2018designing}
A.~B. Henson, P.~S. Gromski, and L.~Cronin, ``Designing algorithms to aid
  discovery by chemical robots,'' \emph{ACS central science}, vol.~4, no.~7,
  pp. 793--804, 2018.

\end{thebibliography}

\end{document}